\documentclass[11pt]{article}

\usepackage{acl}
\usepackage{times}
\usepackage{latexsym}
\usepackage{booktabs}
\usepackage{multirow}
\usepackage{threeparttable}
\usepackage{subfig}

\usepackage[T1]{fontenc}

\usepackage[utf8]{inputenc}

\usepackage{microtype}

\usepackage{inconsolata}

\usepackage{graphicx}
\usepackage{amsmath}
\usepackage{amsfonts}
\usepackage{mathtools}
\usepackage{booktabs}
\DeclareMathOperator*{\argmax}{arg\,max}

\usepackage{bbm}
\usepackage{pifont}
\usepackage{newunicodechar}

\newunicodechar{✅}{\ding{51}}
\newunicodechar{❌}{\ding{55}}
\newcommand{\tokC}{\ding{51}}   
\newcommand{\tokI}{\ding{55}}   

\usepackage[most]{tcolorbox}
\usepackage{ragged2e}   
\usepackage{placeins}

\title{Score $\times$ Decoder: A Unified View of Unsupervised Inference-Time Scaling for Hallucination Mitigation}

\author{%
    Yun-Chen Cheng\thanks{These authors contributed equally to this work.}
    \quad Che-Yu Lin\textsuperscript{\footnotemark[1]}
    \quad Cheng-Lin Yang \\
  CyCraft AI Lab, Taiwan\\
  \texttt{hhhhaura@gmail.com}, \texttt{\{jerry.lin,cl.yang\}@cycraft.com} \\
}

\begin{document}
\maketitle
\begin{abstract}
Large language models hallucinate even when the answer lies within their parameters. While inference-time scaling can surface this latent knowledge, the most effective methods require supervision: a
trained verifier or reward model. We ask what can be done with only a base
language model: which intrinsic signal best identifies correct outputs, and how
should it be decoded? We cast this as a score~$\times$~decoder grid pairing four
scores (perplexity, contrastive, power-distribution likelihood, and
self-verification) with three decoding families (optimization, sampling,
consensus), and evaluate every cell on MATH500 with the base and
instruction-tuned Qwen3-1.7B. While self-verification, which prompts the model to judge its own answer and is sharpened by a training-free virtual-thinking prefix, works well in most settings, no score has a fixed quality: its value depends on the decoder that consumes it and on model capability. When no supervision is available, the score and the decoding family must be chosen together.
\end{abstract}

\section{Introduction}
Large language models hallucinate. However, to mitigate these errors, we must first distinguish between two fundamentally different pathologies: the answers a model \emph{could} have gotten right but didn't, and the ones it \emph{couldn't} have gotten right but won't admit. Within the taxonomy of the know--unknow quadrant~\citep{yin-etal-2023-large, li2024honesty}, these manifest as \emph{unknown knows}, elicitation failures where the correct knowledge is trapped in the model's weights but fails to surface, and \emph{unknown unknowns}, awareness failures where the model lacks the knowledge entirely yet blindly fabricates. While traditionally treated as separate problems evaluated by distinct metrics (pass$@1$ for elicitation versus calibration, abstention, or selective accuracy for awareness), both failure modes share a highly compelling virtue: they can, in principle, be intercepted and corrected at inference time without the prohibitive cost of retraining. Elicitation failures can be rescued by better navigating the generation space through strategic sampling, whereas awareness failures can be neutralized by robustly estimating the model's internal confidence.

We focus in this work on the \emph{unknown knows} subset: elicitation failures, where the correct answer is already latent in the model's weights. The inference-time literature has tools for this (verifier-guided search, learned reward models, best-of-$N$ with trained scorers), but the most effective ones share a dependency: a \emph{supervised} signal, typically a verifier or reward model trained on labeled correctness data. This conflates whether inference-time compute can help with whether a good verifier happens to be available, and restricts deployment to domains where such labels are abundant. We ask the strictly unsupervised version: what can be recovered using only scalar signals extracted from the base LM itself, with no learned verifier, no reward model, no external judge?

The central design choice in this setting is which scalar signal extracted
from the base model best identifies correct outputs, and how that signal
should be composed into a decoding strategy. We evaluate four \emph{gray-box} scoring
functions as parallel candidates: perplexity, contrastive scores
\citep{li2023contrastive, obrien2023contrastive}, power distribution
likelihood, and self-verification \citep{kadavath2022language, xie2023self,
weng2023selfverify} with a virtual thinking prefix, and compose each with three decoding families:
\emph{optimization}, which searches for the highest-scoring candidate
(realized as best-of-$N$); \emph{sampling}, which draws from a
score-shaped target distribution $p_{\mathrm{final}}(y \mid x) \propto
p_{\mathrm{LM}}(y \mid x)\, s(y \mid x)$ (realized as SMC and MCMC);
and \emph{consensus} ($a^\star \in \arg\max_{a} \,\mathbb{E}_{y \sim
p_{\mathrm{final}}}[\mathbf{1}[\mathrm{ans}(y) = a]]$), which selects
the mode of $p_{\mathrm{final}}$ estimated over many samples and is
realized as majority vote \citep{wang2023selfconsistency} and
score-weighted aggregation. This factorization unifies a wide range of
prior inference-time methods into a single score $\times$ decoder grid.

We make three contributions.
\paragraph{(1) Score design.} We identify self-verification with a
\emph{virtual-thinking} prefix as a structurally better-grounded
unsupervised elicitation score than likelihood-derived alternatives. It is \emph{conditioned on the full candidate} and \emph{semantically aligned} with the downstream metric (asking whether an answer is right, not whether it is likely), and is defined as the normalized probability of the correct verdict, $s_{\mathrm{sv}}(y \mid x) = p_{\text{\tokC}}/(p_{\text{\tokC}} + p_{\text{\tokI}})$ (see Section~\ref{sec:score}), making it invariant to the absolute scale of verdict token probabilities.
Self-verification is not new as a signal in isolation~\citep{kadavath2022language, xie2023self}; our contribution is to make it the central score in a unified inference-time framework and to enhance it with virtual thinking.

\paragraph{(2) Virtual thinking without finetuning.} We show that the calibration of self-verification can be sharpened by inserting a sequence of tokens before the verdict logits that consumes additional compute without producing visible reasoning. We find that the choice of prefix matters:
\emph{structured} fillers such as countdowns (\textit{``$50, 49, 48, \dots, 1$''}) outperform \emph{uniform} fillers such as dots (\textit{``$\dots$''}). This departs from prior findings that latent-reasoning filler tokens require training to be useful~\citep{pfau2024lets, goyal2024think}; we argue the gap is structural: verifier-side virtual thinking is easier than generator-side virtual thinking because the verifier only needs to produce a single classification bit, not a coherent token sequence.

\paragraph{(3) Systematic evaluation.} We evaluate every cell of the score $\times$ decoder grid along three complementary axes that correspond to distinct uses of an inference-time score.
\emph{Score quality:} for a fixed pool of candidates, how well does the score rank correct candidates above incorrect ones (local ranking, per-question AUROC and top-$k$ accuracy)?
\emph{Decoding effectiveness:} when combined with a score, do optimization methods (best-of-$N$) or sampling methods (SMC, MCMC) produce more accurate single outputs than vanilla decoding
(pass$@k$, accuracy of the modal sample)?
\emph{Aggregation effectiveness:} when compute permits many samples per question, does score-weighted consensus improve over majority voting or best-of-$N$ selection?
Across these axes, self-verification with virtual thinking performs well on most
evaluations, but no score wins universally. The score~$\times$~decoder
factorization lets us characterize scores and decoders separately, and find that a score's realized quality is not fixed: it depends on the decoder that consumes it, so when no supervision is available the score and the decoding
family should be chosen together.

\paragraph{Scope.}
Our claims concern the interaction structure between scores and decoders,
not absolute rankings across models or domains. Section~\ref{sec:analysis}
and Appendix~\ref{sec:appendix-failure} give mechanistic explanations
(dynamic-range mismatch, particle collapse) that should transfer beyond
our setup, and we leave empirical confirmation at scale to future work.

\section{Related Work}
\paragraph{Score Functions for Inference-Time Guidance.}
Inference-time scoring signals differ in the model access they require;
we focus on \emph{gray-box} signals, which read the model's own
token-level logits rather than only its text. One line of work builds
training-free, unsupervised scores: sequence-level perplexity is the
canonical choice, while \emph{contrastive decoding}~\citep{li2023contrastive,obrien2023contrastive}
contrasts a strong expert's logits against a weaker amateur, recently
extended by \citet{kim2026contrastive} to contrast against a perturbed
thinking trace. Unsupervised \emph{self-verification} instead prompts
the model to judge its own output and reads the verdict logit, e.g.\
$P(\text{True})$~\citep{kadavath2022language,xie2023self}. A second,
\emph{supervised} line trains external estimators: step-level
\emph{process reward models} (PRMs)~\citep{lightman2023lets,wang2024math}
provide learned guidance but require expensive annotation and
specialized training. This raises the question we study: can an unsupervised score read from
the base model alone guide decoding without the annotation cost of a
trained verifier? We vary the scoring function across four gray-box
candidates (perplexity, contrastive scoring, the power distribution, and
self-verification) and map how each interacts with MCMC, consensus
reweighting, and SMC guidance.
 
\paragraph{Sampling-Based Methods.}
Framing generation as probabilistic inference has motivated a line of sampling-based approaches. \citet{lew2023sequential} introduce LLaMPPL, a Sequential Monte Carlo (SMC) framework for controlled generation, later extended with learned twist functions that estimate expected future reward~\citep{zhao2024probabilistic} and to syntactic/semantic constraints~\citep{loula2025syntactic}. \citet{puri2025probabilistic} apply particle-based Monte Carlo to inference-time scaling, reporting 4--16$\times$ better scaling than deterministic search but relying on a process reward model. On the MCMC side, \citet{faria2024quest} propose a Metropolis--Hastings sampler driven by expected reward for translation. A related line of work biases decoding toward a target distribution without resampling at the sequence level, e.g., gradient-based controlled decoding in discrete token space~\citep{pynadath2025controlled}. Most closely related, the concurrent work of \citet{karan2025reasoning} and \citet{ji2026scalable} explores MCMC over a sharpened \emph{power distribution} of the base model. Both papers fix this single scoring choice and report strong gains \emph{in the MCMC setting alone}.
 
\paragraph{Optimization-Based Methods.}
Search-based generation methods cast inference as an optimization problem over a structured token space. The baseline approach, \emph{best-of-$N$}, samples multiple candidates and selects the highest-scoring sequence under a verifier, with recent refinements focusing on early-decoding signals to minimize sampling costs~\citep{wang2025sampling}. More complex frameworks like \emph{Tree-of-Thoughts}~\citep{yao2023tree} and \emph{RAP}~\citep{hao2023reasoning} extend this concept by employing Monte Carlo Tree Search (MCTS) as a policy-improvement operator over tree-structured reasoning paths. This test-time compute scaling paradigm underpins modern frontier models such as OpenAI's o1~\citep{openai2024o1} and DeepSeek-R1~\citep{guo2025deepseekr1}.

 
\paragraph{Consensus-Based Methods.}
Rather than selecting a single candidate sequence, consensus methods aggregate across multiple decoded samples. Standard approaches like \emph{Self-consistency}~\citep{wang2023selfconsistency} rely on a simple majority vote among extractable answers, while weighted variants scale votes by confidence signals~\citep{kang2025scalable} or intermediate path features~\citep{wan2024reasoning}. We incorporate consensus aggregation into our study not to propose a novel voting mechanism, but to serve as a diagnostic lens for evaluating scoring functions. By applying a standard post-hoc reweighting using calibrated probabilities derived from each score, we complete the picture of inference-time scaling behavior.
 
\paragraph{Latent and ``Virtual'' Thinking.} While the above methods differ in how they aggregate or select outputs, they share a reliance on explicit token-level reasoning as thinking; a separate line of work asks whether such thinking can instead be internalized into latent computation.
A parallel line of work attempts to internalize the computational benefits of a chain-of-thought into hidden states or special tokens rather than explicit text. Architectural interventions such as pause-tokens~\citep{goyal2024think}, meta-tokens~\citep{zelikman2024quietstar}, and continuous-thought decoding (Coconut;~\citealp{hao2024training}) all operate in continuous latent spaces. However, current literature paints a pessimistic view of these "virtual" thinking paradigms when applied without heavy training; for instance, \citet{pfau2024lets} demonstrate that LLMs fail to benefit from virtual filler tokens unless they undergo dense, highly specific supervision, leading to a consensus that virtual thinking does not work out-of-the-box.

\section{Score Function Design}
\label{sec:score}
A central ingredient of our framework is a scalar score $s(y \mid x) \in \mathbb{R}^{+}$ that measures the correctness of a response $y$ to a query $x$. Rather than relying on an external verifier or a separately trained reward model, we design a score that is \emph{intrinsic}: it is computed by prompting the same language model that generated $y$ to evaluate its own
output. Prior work on virtual reasoning has focused on the generator side, where the model must produce coherent token sequences, a task for which plug-and-play filler tokens have been shown to be ineffective without dense supervision~\citep{goyal2024think, pfau2024lets}. As we describe below, we instead repurpose virtual thinking as a \emph{scoring-side device}, a structurally simpler use case that succeeds completely plug-and-play. This section describes the two components of the score: the self-verification prompt and the virtual thinking countdown, and defines the resulting scoring function.
\subsection{Self-Verification Score}
\label{sec:score-prompt}
Given a query $x$ and a candidate response $y$, we construct a verification prompt $v(x, y)$ that instructs the language model to judge whether $y$ correctly solves $x$.
The prompt is designed so that the model's first generated token is one of two special verdict tokens, \tokC\ (correct) or \tokI\ (incorrect),%
\footnote{The actual tokens used are the Unicode characters ✅ (\texttt{U+2705}) and ❌ (\texttt{U+274C}); \LaTeX\ renders them here via \texttt{\textbackslash ding\{51\}} and \texttt{\textbackslash ding\{55\}} as visual approximations.}
followed optionally by a brief explanation; see Appendix~\ref{app:prompts} for the full template and grading rules.
By terminating the prompt with the incomplete string \texttt{"\#\#\# Verdict: "}, we arrange for the model's next token to be either \tokC\ or \tokI, whose log-probabilities are directly readable from the output distribution.
Let $p_{\text{\tokC}}(x, y)$ and $p_{\text{\tokI}}(x, y)$ denote the probabilities that the language model assigns to the verdict tokens \tokC\ and \tokI, respectively, as the next token following $v(x, y)$.
We define the \emph{self-verification score} as the odds ratio of the correct verdict:%
\footnote{This score is well-defined provided the model assigns negligible probability to tokens other than \tokC\ and \tokI\ at the verdict position. We empirically verify that $p_{\text{\tokC}}(x,y) + p_{\text{\tokI}}(x,y) \approx 1$ holds across inputs in Appendix~\ref{app:score-validation}, confirming that the verification prompt reliably elicits the intended verdict format.}
\begin{equation}
    s_{\mathrm{sv}}(y \mid x) \;=\; \frac{p_{\text{\tokC}}(x, y)}{p_{\text{\tokI}}(x, y)}.
    \label{eq:score}
\end{equation}
\subsection{Virtual Thinking Countdown}
\label{sec:score-countdown}
Prompting a model for an immediate verdict is often unreliable because it precludes prior reasoning. To encourage implicit deliberation without triggering an explicit chain-of-thought, we insert a \emph{virtual thinking prompt} before the \texttt{"\#\#\# Verdict:"} line (see Appendix~\ref{app:prompt-selfverify}). We evaluate two variants: a \textbf{countdown filler} (\texttt{$n, n{-}1, \ldots, 1$}) and a \textbf{dot filler} ($d$ repeated dots). Both allocate input tokens for implicit reasoning before forcing a verdict. Because this filler is part of the input context, $s_{\mathrm{sv}}$ maintains its single-forward-pass evaluation cost regardless of length. We compare and ablate these variants in Appendix~\ref{app:virtual_ablation}.

\section{Elicitation Framework}
\label{sec:elicitation}
Given a scoring function $s(y \mid x)\in\mathbb{R}^+$ that evaluates the quality of a response $y$ to a query $x$
, a natural question is how to best spend additional inference-time compute to improve output quality.
We organize the answer around two complementary strategies.
The first, \emph{optimization} (Section~\ref{sec:optimization}), uses the score to search for a single high-quality output.
The second, \emph{consensus} (Section~\ref{sec:consensus}), instead aggregates over a collection of candidate outputs drawn from a score-shaped distribution, trading the commitment of picking one best candidate for the stability of collective agreement.
Both strategies are applicable to any scoring function, and can in principle be combined.
\subsection{Optimization}
\label{sec:optimization}
The most direct use of a score is to treat inference as a search problem: generate candidate responses and return the one with the highest score.
The simplest instantiation is best-of-$N$, which draws $N$ responses $y_1, \dots, y_N \sim p_{\mathrm{LM}}(\cdot \mid x)$
\footnote{Throughout this paper, $p_{\mathrm{LM}}(y \mid x)$ denotes the probability of generating $y$ given input $x$ under the language model, where $x$ is formatted according to the prompt template described in Appendix~\ref{app:prompts}.}
independently from the language model and returns
\begin{equation}
    {y}^* = \argmax_{i \in [n]}\; s(y_i \mid x).
    \label{eq:best-of-n}
\end{equation}
{More sophisticated search algorithms such as beam search and MCTS can exploit the sequential structure of generation to allocate compute more efficiently, and we leave a thorough empirical comparison to future work.}
\subsection{Sampling from the Score-Shaped Distribution}
\label{sec:sampling}
Rather than optimizing for a single output, we can define a \emph{score-shaped} target distribution
\begin{equation}
    p_{\mathrm{final}}(y \mid x) \;\propto\; p_{\mathrm{LM}}(y \mid x)\cdot s(y \mid x),
    \label{eq:pfinal}
\end{equation}
which up-weights responses that are both likely under the language model and assigned high score.
The inclusion of $p_{\mathrm{LM}}$ in the target (rather than defining $p_{\mathrm{final}} \propto s$ alone) is a principled design choice whose necessity we discuss in Appendix~\ref{sec:appendix-failure}.
Sampling from $p_{\mathrm{final}}$ is non-trivial because the normalizing constant is intractable and the support is the space of all token sequences. In practice, this constant is handled implicitly: in MCMC, it cancels in the Metropolis--Hastings acceptance ratio (Eq. \ref{eq:score_acceptance_rate}); in SMC, the resampling step normalizes particle weights at each step, so no explicit computation of the constant is required.
\paragraph{Markov Chain Monte Carlo (MCMC).}
MCMC methods construct a Markov chain whose stationary distribution is $p_{\mathrm{final}}$.
Concretely, we adopt the \emph{prefix-resample} proposal of \citet{karan2025reasoning}, in which a new candidate is generated by sampling a random prefix of the current response and then continuing it autoregressively; we omit the blockwise extension described in that work and refer readers there for the full procedure.
The resulting Metropolis--Hastings chain accepts a proposed response $y'$ with probability
\begin{equation}
    \label{eq:score_acceptance_rate}
    \alpha = \min\!\left(1,\; \frac{s(y' \mid x)}{s(y \mid x)}\right),
\end{equation}
ensuring detailed balance with respect to $p_{\mathrm{final}}$, by specialization of the general Metropolis--Hastings acceptance rule (see Eq.~\ref{eq:acceptance_rate}).
After a suitable burn-in period, the chain produces approximate samples from the target distribution.
\paragraph{Sequential Monte Carlo (SMC).}
Following the basic SMC formulation of \citet{Doucet2001}, SMC methods maintain a population of $N$ weighted \emph{particles} $\{y_{1:t}^{(i)}, w_t^{(i)}\}_{i=1}^{N}$ that are propagated token-by-token through the generation process.
At each step $t$, the language model extends each particle by one token under the proposal distribution, and the particle weight is updated as
\begin{equation}
    w_t^{(i)} \;\propto\; s(y_{1:t}^{(i)} \mid x),
    \label{eq:smc-weight}
\end{equation}
where $s$ is evaluated on the partial sequence $y_{1:t}^{(i)}$.
Periodically, a resampling step duplicates high-weight particles and discards low-weight ones, concentrating compute on promising prefixes.
At termination, each surviving particle is an approximate draw from $p_{\mathrm{final}}$. \footnote{The formulation requires the score to be applicable to \emph{partial} sequences. Alternatively, if the score is only meaningful for complete sequences, the weight of a partial sequence $y_{1:t}$ can instead be set to the expected score over possible completions, $w_t \;\propto\; \mathbb{E}\bigl[s(y \mid x) \mid y_{1:t}\bigr]$, which can be approximated by averaging the scores of several sampled completions of $y_{1:t}$. We leave this formulation to future work and use partial-sequence scoring in our experiments.}
\paragraph{Remark: two roles of $p_{\mathrm{LM}}$.}
$p_{\mathrm{LM}}$ is used in two ways here, and the two are configured
independently. As the \emph{proposal}, it draws candidate sequences,
where we are free to use practical settings (temperature $\tau<1$,
top-$k$, top-$p$) for fluent, diverse candidates. As the basis of the
\emph{score}, it supplies the probabilities that $s(y\mid x)$ reads;
this is always evaluated at $\tau=1$ with no truncation, reflecting the
Conflating the two would distort $s$ and bias the sampler
away from the intended $p_{\mathrm{final}}$.
Proposal hyperparameters are in Section~\ref{sec:experiments}.
\subsection{Consensus}
\label{sec:consensus}
Given a set of candidates, consensus methods select the response that is most representative of the ensemble rather than the one with the highest individual score.
Formally, we seek
\begin{equation}
    a^{*} = \argmax_{{a}}\; \mathbb{E}_{y \sim p_{\mathrm{final}}}\bigl[\mathbf{1}[\mathrm{ans}(y) = {a}]\bigr],
    \label{eq:consensus-obj}
\end{equation}
i.e., the mode of $p_{\mathrm{final}}$.
This objective can be estimated in two ways depending on how the candidates were obtained.
\paragraph{Majority vote for MCMC/SMC samples.}
When candidates $y_1, \dots, y_n$ are (approximate) draws from $p_{\mathrm{final}}$, as produced by the MCMC or SMC samplers of Section~\ref{sec:sampling}, the expectation in Eq.~\eqref{eq:consensus-obj} can be estimated directly by unweighted majority vote:
\begin{equation}
    a^{*} \approx \argmax_{{a}}\; \frac{1}{n}\sum_{i=1}^{n} \mathbf{1}[\mathrm{ans}(y_i) = {a}].
\end{equation}
No reweighting is needed because the sampling distribution already matches the target.
\paragraph{Importance-weighted vote for standard LM samples.}
When candidates are drawn from the base model $p_{\mathrm{LM}}$ rather than from $p_{\mathrm{final}}$, a direct majority vote is biased toward regions of high $p_{\mathrm{LM}}$ that may not be favored by the score.
We correct for this mismatch via importance weighting.
By a change of measure,
\begin{align}
    &\mathbb{E}_{y \sim p_{\mathrm{final}}}\bigl[\mathbf{1}[\mathrm{ans}(y) = {a}]\bigr]\\
    &= \mathbb{E}_{y \sim p_{\mathrm{LM}}}\!\left[\frac{p_{\mathrm{final}}(y \mid x)}{p_{\mathrm{LM}}(y \mid x)}\,\mathbf{1}[\mathrm{ans}(y) = {a}]\right] \notag\\
    &\propto \mathbb{E}_{y \sim p_{\mathrm{LM}}}\bigl[s(y \mid x)\,\mathbf{1}[\mathrm{ans}(y) = {a}]\bigr],
    \label{eq:iw-vote}
\end{align}
where the second line uses Eq.~\eqref{eq:pfinal} and drops the normalizing constant, which is the same for all ${a}$ and does not affect the $\argmax$.
This yields the weighted consensus estimator
\begin{equation}
    a^{*} \approx \argmax_{{a}}\; \sum_{i=1}^{n} s(y_i \mid x)\,\mathbf{1}[\mathrm{ans}(y_i) = {a}],
    \label{eq:weighted-mv}
\end{equation}
which reduces to standard majority vote when all scores are equal and to best-of-$N$ when candidates are near-unique.

\section{Experiments}
\label{sec:experiments}

\subsection{Settings}
\label{sec:exp-settings}
\paragraph{Dataset.}
We evaluate on \textbf{MATH500}~\citep{lightman2023lets}, a 500-problem subset of the MATH benchmark~\citep{hendrycks2021math}. It is the smallest widely-used benchmark that simultaneously offers verifiable correctness, extractable answers, and a sufficient difficulty range: the properties our score~$\times$~decoder analysis requires. We score exact-match on the final boxed answer after normalizing whitespace and symbolic forms.

\paragraph{Models.}
We use the \textbf{Base} (pretrained) and \textbf{Chat}
(instruction-tuned) variants of \textbf{Qwen3-1.7B}~\citep{qwen3technicalreport}.
Drawing both from one model family fixes architecture, tokenizer, and pretraining, so any Base-vs-Chat difference isolates the effect of instruction tuning. Each variant is its own generator and scorer (no external judge), and all runs use \texttt{bfloat16}. We fix model scale to keep the score~$\times$~decoder grid well-populated, since partial coverage at larger scales would not support our interaction claims.

\paragraph{Generation budget and token lengths.}
All sampling uses $\tau = 0.7$, top-$k = 50$, and top-$p = 0.95$. Chat candidates are 2{,}048-token continuations following a 2{,}048-token truncated thinking trace; Base candidates are 4{,}096-token completions. All post-hoc methods (ranking, optimization, best-of-$N$, reweighted consensus) share a pool of $N = 100$ candidates per question. SMC runs $N = 100$ particles, bootstrapping every 50 tokens. MCMC runs 100 independent Metropolis--Hastings chains of 10 steps each, retaining the final state of each as one sample.

\subsection{Score Functions}
In all score definitions below, $p_{\mathrm{LM}}$ is evaluated at $\tau = 1$ with no top-$k$ or top-$p$ truncation, consistent with the remark in Section~\ref{sec:sampling}.

\paragraph{Perplexity.}
$s_{\mathrm{ppl}}(y \mid x) = p_{\mathrm{LM}}(y \mid x)^{1/|y|}$.
The geometric mean of token probabilities, inversely related to sequence perplexity.

\paragraph{Power distribution score.}
$s_\beta(y \mid x) = p_{\mathrm{LM}}(y \mid x)^\beta$.
The base-model distribution sharpened to a power $\beta > 1$.
We adopt $\beta = 4.0$ from \citet{karan2025reasoning} but retain our standard generation temperature of $0.7$ (rather than the recommended $1/\beta = 0.25$) to keep sampling conditions consistent across all scoring methods.

\paragraph{Contrastive score.}
$s_{\mathrm{CD}}(y \mid x) = p_{\text{EXP}}(y \mid x) /p_{\text{AMA}}(y \mid x)$.
The ratio of expert to amateur likelihoods, amplifying tokens uniquely favored by deliberative reasoning~\citep{li2023contrastive}. \footnote{For both Chat and Base model, $p_\mathrm{AMA}=p_\mathrm{LM}$. For the Chat model, $p_\mathrm{EXP}$ is $p_\mathrm{LM}$ with a thinking trace in the prompt. For the Base model, the $p_\mathrm{EXP}$ is $p_\mathrm{LM}$ with a structured analysis prefix (generated with the prompt \ref{app:prompt-contrastive}; truncated to 2{,}048 tokens) in the prompt.} In the SMC condition only, we use the length-normalized variant $s_{\mathrm{CD}}(y \mid x)^{1/|y|}$ to prevent length bias from eliminating already-completed particles.

\paragraph{Self-verification score.}
($s_{\mathrm{sv}}$, Eq.~\eqref{eq:score}, Section~\ref{sec:score}). We report three variants: a no-virtual-thinking baseline, a \textbf{dot filler} with $d=50$, and a \textbf{countdown filler} with $n=50$. The complete ablation over filler type and length is reported in Appendix~\ref{app:virtual_ablation}.

\subsection{Metrics}
\label{sec:exp-metrics}
We evaluate scores along three axes: candidate \emph{ranking}, distribution \emph{shaping} via score-guided sampling, and \emph{consensus} improvement.
Throughout, $x$ denotes a question from set $\mathcal{X}$,
Normal/SMC/MCMC pool (of $x$) denotes candidates $\{y_1,\dots,y_N\}$ drawn from the corresponding sampling methods.

\paragraph{Ranking.}
We report \textbf{Top-$k$ accuracy}: the fraction of questions for which at least one correct candidate appears among the $k$ highest-scored candidates ($k{=}1$ recovers best-of-$N$ accuracy; $k{=}N$ recovers oracle pass@$N$), and \textbf{per-question AUROC}: the probability that a randomly drawn correct candidate outscores a randomly drawn incorrect one, treating $s(\cdot\mid x)$ as a binary classifier within each question. We report the CDF of AUROC across questions containing at least one correct and one incorrect candidate. Both metrics are evaluated on the Normal pool.

\paragraph{Sampling.}
We report \textbf{pass@$k$}: the probability that at least one of $k$ i.i.d.\ draws from the corresponding pool is correct, computed via the standard unbiased estimator~\citep{chen2021codex} and averaged over $\mathcal{X}$.\footnote{For Normal decoding and MCMC the $k$ samples are independent---each MCMC output is the final state of a separate chain---whereas SMC particles are correlated through shared ancestry via resampling, so pass@$k$ for SMC should be read with this caveat.} We also report mean single-sample accuracy and mean number of distinct answers per question, characterizing the diversity of the distribution induced by each decoding method.

\paragraph{Consensus.}
We report \textbf{consensus accuracy}: the fraction of questions whose aggregated answer is correct, under the methods of Section~\ref{sec:consensus}.

\section{Results}
\label{sec:results}
\begin{figure*}[!ht]
    \centering
    \includegraphics[width=\textwidth]{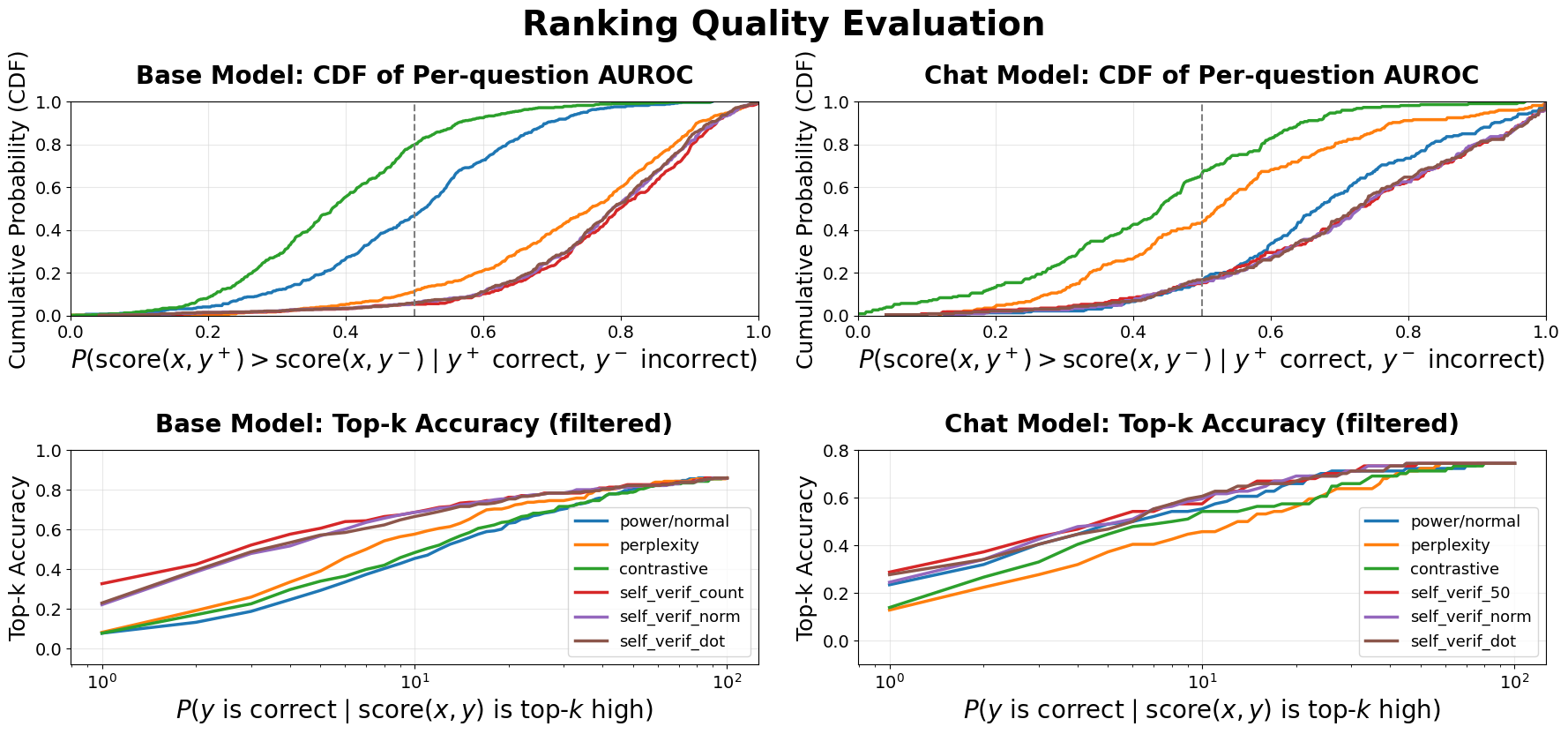}
    \caption{\textbf{Ranking quality of different scores} for the Base (left) and Chat (right) models. \emph{Top:} CDF of per-question AUROC over questions with valid pools; mass to the right of the $0.5$ line indicates above-chance discrimination (a curve shifted rightward is better). \emph{Bottom:} Top-$k$ accuracy on the hard subset (oracle accuracy ${\leq}0.5$).
    }
    \label{fig:ranking}
\end{figure*}
\subsection{Ranking Ability}
\label{sec:result-rank}
Self-verification ranks highest on nearly every metric: it achieves the best per-question AUROC on both models and recovers the most correct answers at $k{=}1$ on the hard subset (Figure~\ref{fig:ranking}). Power scores competitively on Chat but poorly on Base; its apparent strength on Chat reflects the model's already-concentrated output distribution rather than genuine discriminative ability. Perplexity offers weak ranking signal on Base and essentially none on Chat. Contrastive is the weakest score throughout; on Base it relies on an engineered analysis prefix as a substitute for the intrinsic expert/amateur pair available on Chat (a thinking-trace vs.\ bare-prompt continuation), which explains why contrastive underperforms more severely on Base than the other scores.

On the Chat model, power and self-verification are nearly indistinguishable under Top-$k$ accuracy. Figure~\ref{fig:ranking} therefore reports Top-$k$ results on the hard subset (questions with oracle accuracy ${\leq}0.5$) to expose score differences; the unfiltered version is given in Figure~\ref{fig:ranking_quality_full}, Appendix~\ref{app:full_figures}.

\subsection{Sampling Quality}
\label{sec:result-sample}
\begin{figure*}[!tp]
    \centering
    \includegraphics[width=\textwidth]{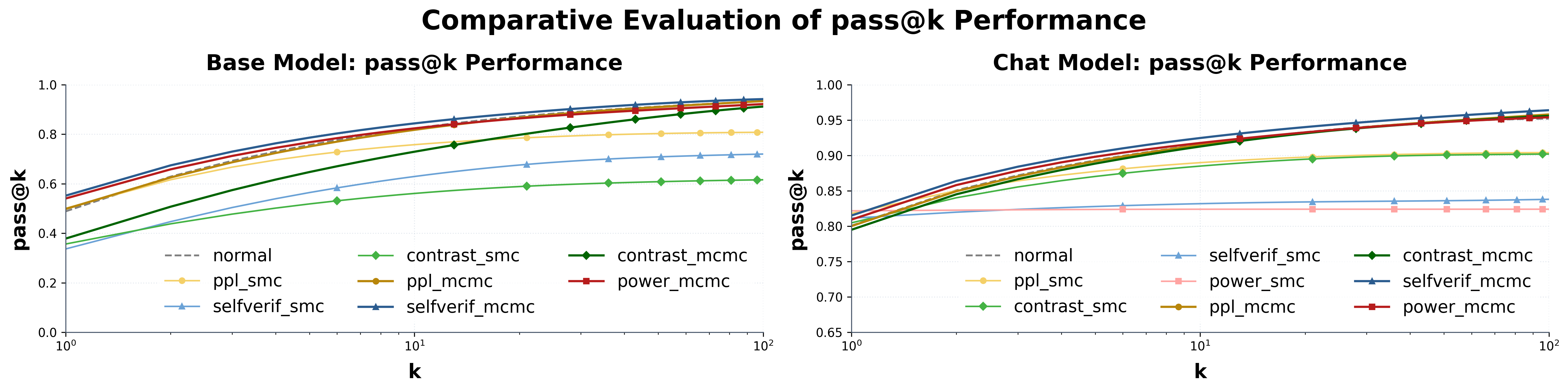}
    \caption{\textbf{pass@$k$ by decoding method}, on the Base (left) and Chat (right) models, averaged over $\mathcal{X}$.
    }
    \label{fig:passk}
\end{figure*}

\begin{table}[ht]
\centering
\small
\setlength{\tabcolsep}{4pt}
\begin{threeparttable}
\begin{tabular}{lcccc}
\toprule
& \multicolumn{2}{c}{\textbf{Single-sample Acc.}} & \multicolumn{2}{c}{\textbf{\#Distinct Ans.}} \\
\cmidrule(lr){2-3} \cmidrule(lr){4-5}
\textbf{Decoding} & \textbf{Base} & \textbf{Chat} & \textbf{Base} & \textbf{Chat} \\
\midrule
Normal            & 0.488 & 0.801 & 97.63 & 99.98 \\
\midrule
PPL(SMC)         & 0.501 & 0.818 & 17.75 & 26.46 \\
Self-Verif(SMC)  & 0.337 & 0.811 & 13.30 & 8.20 \\
Contrast(SMC)    & 0.356 & 0.805 & 17.33 & 30.00 \\
Power(SMC)$^*$   & --    & \textbf{0.822} & --    &  1.55 \\
\midrule
PPL(MCMC)        & 0.498 & 0.800 & 99.99 & 99.99 \\
Self-Verif(MCMC) & \textbf{0.552} & 0.815 & 99.99 & 99.99 \\
Contrast(MCMC)   & 0.379 & 0.803 & 100.0 & 99.99 \\
Power(MCMC)      & 0.540 & 0.809 & 99.87 & 99.75 \\
\bottomrule
\end{tabular}
\begin{tablenotes}
\small
\item[$^*$] Excluded for the Base model: produces degenerate non-stopping sequences under this setting.
\end{tablenotes}
\end{threeparttable}
\caption{Mean single-sample accuracy and number of distinct answers per question by decoding method.}
\label{tab:meanacc}
\end{table}
MCMC variants track or exceed unguided decoding across the full pass@$k$ curve, while SMC variants saturate early (Figure~\ref{fig:passk}). The divergence traces to SMC's resampling: permanently concentrating weight onto a shrinking set of prefixes collapses diversity to $8$--$30$ distinct answers per question (just $1.55$ for power-guided SMC), compared to ${\approx}100$ for MCMC and unguided decoding (Table~\ref{tab:meanacc}). SMC stakes the outcome on a few high-scoring prefixes while discarding the coverage that makes sampling valuable; when the score is unreliable, no diversity remains to recover. Self-verification paired with MCMC yields the strongest distribution on both models: best Base single-sample accuracy and best Chat pass@$k$.

\subsection{Consensus Quality}
\label{sec:result-consensus}
\begin{figure}[!ht]
    \centering
    \includegraphics[width=\columnwidth]{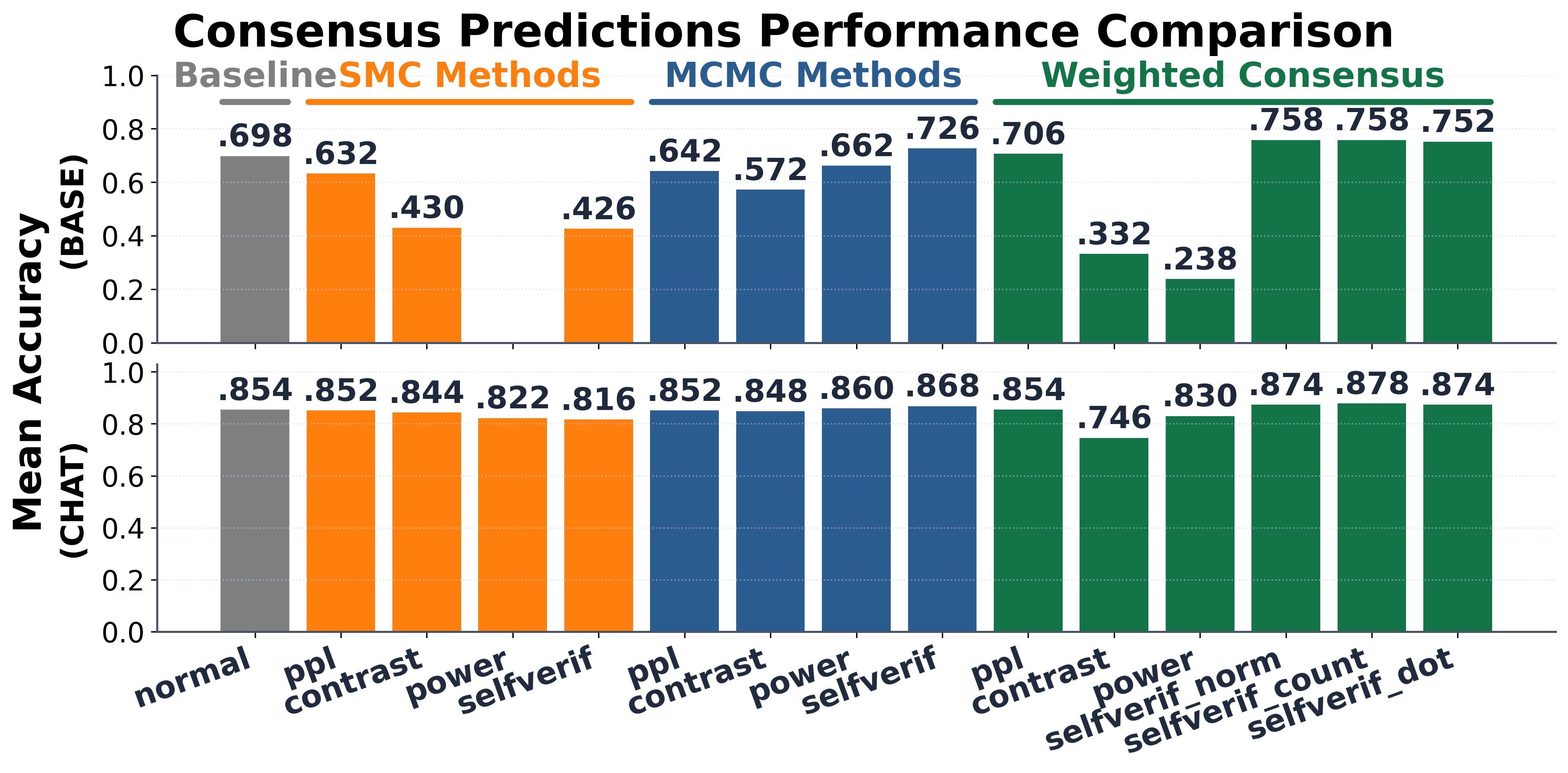}
    \caption{\textbf{Consensus accuracy}
    for Base (top) and Chat (bottom) models, grouped by the aggregation method (baseline is majority vote over Normal pool).}
    \label{fig:consensus}
\end{figure}

Self-verification is the most reliable consensus weight: reweighting the Normal pool by $s_{\mathrm{sv}}$ yields the highest weighted-consensus accuracy, and it is also the best MCMC consensus guide (Figure~\ref{fig:consensus}). The likelihood-derived scores are less stable: the large dynamic range of power and perplexity lets a single high-score candidate dominate the weighted vote and degenerate toward best-of-$N$, most visibly on the Base model where power-reweighted consensus collapses. SMC-based consensus is competitive only when the SMC pool has not collapsed onto a few incorrect modes (Table~\ref{tab:meanacc}).
Virtual thinking does not improve consensus: self-verification with and without fillers are near-indistinguishable in consensus accuracy, suggesting its benefit is specific to ranking.

\section{Analysis}\label{sec:analysis}
\paragraph{Decoding choice matters more on Base model.}
Score-guided methods yield larger gains over unguided decoding on Base than on Chat, plausibly because instruction tuning already concentrates the Chat model's output distribution toward correct answers, leaving less room for any score or decoder to improve upon and compressing observable differences between score functions.

\paragraph{Virtual thinking sharpens top-$k$ only.}
Virtual thinking moves only Top-$k$: per-question AUROC and consensus accuracy show no significant difference across variants. At small $k$, however, the variants separate clearly (an ${\approx}9$-point gap on Base and ${\approx}3$-point on Chat at $k{=}1$) before converging by $k{\approx}15$--$30$. Virtual thinking thus makes the single best candidate more reliably first without improving broader ordering. The optimal filler type is model-dependent: countdown leads at small $k$ on Base, while variants converge on Chat, likely because Chat's reasoning traces already exhibit frequent self-checking behavior, reducing the marginal benefit of explicit countdown cues.

\paragraph{Reweighting a vote is not the same as guiding a sampler.}
Reweighted consensus aggregates raw score magnitudes, so a single high-scoring outlier can hijack the vote and degrade the weighted vote toward best-of-$N$ behavior. Samplers such as MCMC, by contrast, consume scores probabilistically against competing candidates, preserving diversity rather than surrendering to extremes. This asymmetry explains why a peaky score like power fails under reweighted consensus (where outlier domination goes unchecked) yet succeeds under MCMC, where no single candidate can monopolize selection. Ultimately, a score's value in reweighting is bounded by its best-of-$N$ quality; its value in sampling is not.

\section{Conclusion}
\label{sec:conclusion}

We studied inference-time scaling for unknown knows, where the correct answer is latent in the model's weights but fails to surface, in the fully unsupervised setting: a base language model with no learned verifier, reward model, or external judge. We evaluated four unsupervised gray-box scores across optimization, sampling, and consensus decoders. Self-verification with a virtual-thinking prefix performs best on most evaluations: without finetuning, it ranks candidates reliably, guides MCMC most effectively, and is the most dependable consensus weight. Still, no score has fixed quality: its realized value depends on the decoder that consumes it and on model capability. Concretely, self-verification with MCMC is robust across both models, whereas SMC is fragile and score-weighted consensus needs a stable score. When no supervision is available, the score and decoding family must therefore be chosen together.

\section*{Limitations}
\label{sec:limitations}
\paragraph{Scope and transferability.}
All experiments use Qwen3-1.7B (Base and Chat) on MATH500, and our conclusions should be read within those boundaries. On the model side, scale may shift the picture unevenly: self-verification could improve as larger models tend to be better calibrated, while likelihood-based scores may weaken against a stronger base distribution, so whether score and decoder rankings hold in the 7B--70B regime or under different RLHF recipes remains open. On the domain side, MATH500's verifiable, extractable answers are more favorable to self-verification than open-ended tasks would be, since the verification prompt can straightforwardly ask about format compliance and numerical correctness; whether SV+VT offers comparable advantages on reading comprehension, summarization, or code generation remains to be established. The virtual-thinking findings carry an additional transferability caveat: the relative effectiveness of countdown versus dot fillers is likely sensitive to the model's pretraining distribution and tokenization scheme, and a filler pattern that aligns with frequent subword patterns in Qwen3's vocabulary may not transfer to models with different vocabularies. We leave cross-scale, cross-domain, and cross-tokenizer confirmation to future work.

\paragraph{Compute overhead of self-verification.}
Self-verification with virtual thinking requires an additional forward
pass per candidate to obtain verdict logits, effectively doubling
the per-candidate inference cost relative to likelihood-based scores.
In our $N{=}100$ pool setting this overhead is absorbed into a
fixed compute budget, but it becomes non-trivial in interactive or
latency-sensitive settings.
We do not report wall-clock runtimes or token-throughput
comparisons, and we do not explore whether a cheaper approximation
(e.g., a single shared verification pass, or early exit on the
filler sequence) can preserve most of the quality gain.

\paragraph{Domain scope of virtual thinking.}
Our finding that structured countdowns outperform uniform fillers as virtual-thinking prefixes was established on mathematical reasoning benchmarks. It is plausible that the advantage is domain-specific: countdown sequences may prime the model toward the sequential, stepwise computation that arithmetic problems reward, and may confer no analogous benefit (or may even harm calibration) on tasks requiring open-ended retrieval, commonsense reasoning, or long-form generation. We leave the design of domain-appropriate virtual-thinking prefixes to future work.

\paragraph{Optimization-based decoding not yet systematically evaluated.}
Our framework positions score functions as general-purpose guides for a family of optimization-based decoders (including best-of-$N$ selection, beam search, and tree-search methods such as MCTS), and the scores are defined and motivated in this broader context. However, we do not yet report systematic experiments across these strategies. The ranking metrics in Section~\ref{sec:exp-metrics} (top-$k$ accuracy, AUROC) capture a slice of this picture (top-$k$ at $k{=}1$ recovers best-of-$N$ accuracy on a fixed pool), but they do not vary the candidate budget, explore structured search over partial sequences, or compare tree-shaped vs.\ flat search at matched compute. A complete evaluation would benchmark score-guided beam search and MCTS against unguided sampling across a range of compute budgets and characterize the conditions under which tree-structured optimization outperforms sampling-based alternatives. We leave this to future work.


\paragraph{SMC with fixed resampling schedule.}
Our SMC implementation uses a fixed, uniform resampling schedule and does not explore variance-reducing alternatives such as learned twist functions or adaptive particle budgets. The SMC results should therefore be interpreted as characterizing naive importance-weighted sampling under each score, not as an upper bound on SMC-based inference.

\paragraph{Fixed MCMC hyperparameters.}
Our MCMC experiments use a fixed burn-in and chain length of 10 steps across all benchmarks and scores, without ablating sensitivity to these choices. Longer chains or adaptive schedules may yield different results.

\paragraph{Token-level guided decoding not evaluated.}
Our framework treats scores as sequence-level functionals applied to complete or partial candidates, composed post-hoc with samplers and optimizers.
A complementary family of methods intervenes instead at the token level, biasing the next-token distribution at each decoding step, including the original autoregressive formulation of contrastive decoding \citep{li2023contrastive} and the more recent contrastive thinking decoding \citep{kim2026contrastive}, which contrasts a perturbed thinking trace against a fluent one.
Because these methods modify the generation process itself rather than reranking completed sequences, they sit outside the score~$\times$~decoder factorization we study.
A direct comparison (in both output quality and compute budget) between token-level guided decoding and our sequence-level framework remains an open question.

\paragraph{Potential risks.}
While this work targets hallucination \emph{reduction}, one
downstream risk merits acknowledgment.
The inference-time scaling methods we study substantially
increase compute requirements relative to standard decoding,
contributing to higher energy consumption; this overhead may also
widen the gap between well-resourced and resource-constrained
practitioners.
We caution against deploying these methods in resource-constrained
settings without careful consideration of computational cost.



\bibliography{custom}

\appendix
\section{Ethics Statement}
\label{app:ethics}

\paragraph{Artifact use.}
This work uses two existing artifacts: the MATH500 benchmark~\citep{lightman2023lets} (a 500-problem research evaluation subset of MATH~\citep{hendrycks2021math}) and the Qwen3-1.7B model family~\citep{qwen3technicalreport}.
Both are used strictly for research evaluation, consistent with their intended purpose.
MATH500 is a publicly released research benchmark; our use is limited to model evaluation and we do not redistribute it.
Qwen3-1.7B is released under the Apache 2.0 license, which permits research use.
We do not use either artifact for any commercial purpose, and any derivatives produced in this work (e.g., score logs, ranked candidate sets) are intended solely for academic research.

\paragraph{Compute and environmental impact.}
All experiments use a single small model (Qwen3-1.7B in \texttt{bfloat16}) evaluated on a 500-question benchmark, representing a modest computational footprint. We do not train or fine-tune any models.

\paragraph{Potential harms.}
Our work studies scoring functions for ranking model-generated mathematical solutions. We do not collect human data, handle personally identifiable information, or deploy systems in consequential real-world settings. The methods could in principle be misused to automate test-taking, but the scope of this paper is limited to research on mathematical reasoning benchmarks.

\paragraph{Intended use of released artifacts.}
Any code or data released alongside this paper is intended for academic research on language model evaluation and mathematical reasoning. It should not be used outside of research contexts.

\paragraph{Use of AI assistants.}
The authors used AI writing assistants (including large language models) to assist with drafting and editing portions of this paper, and AI coding assistants to support code development. All content was reviewed, verified, and revised by the authors, who take full responsibility for the accuracy of the paper.

\section{Full Ranking Quality Figures}
\label{app:full_figures}

\begin{figure*}[t]
    \centering
    \includegraphics[width=\linewidth]{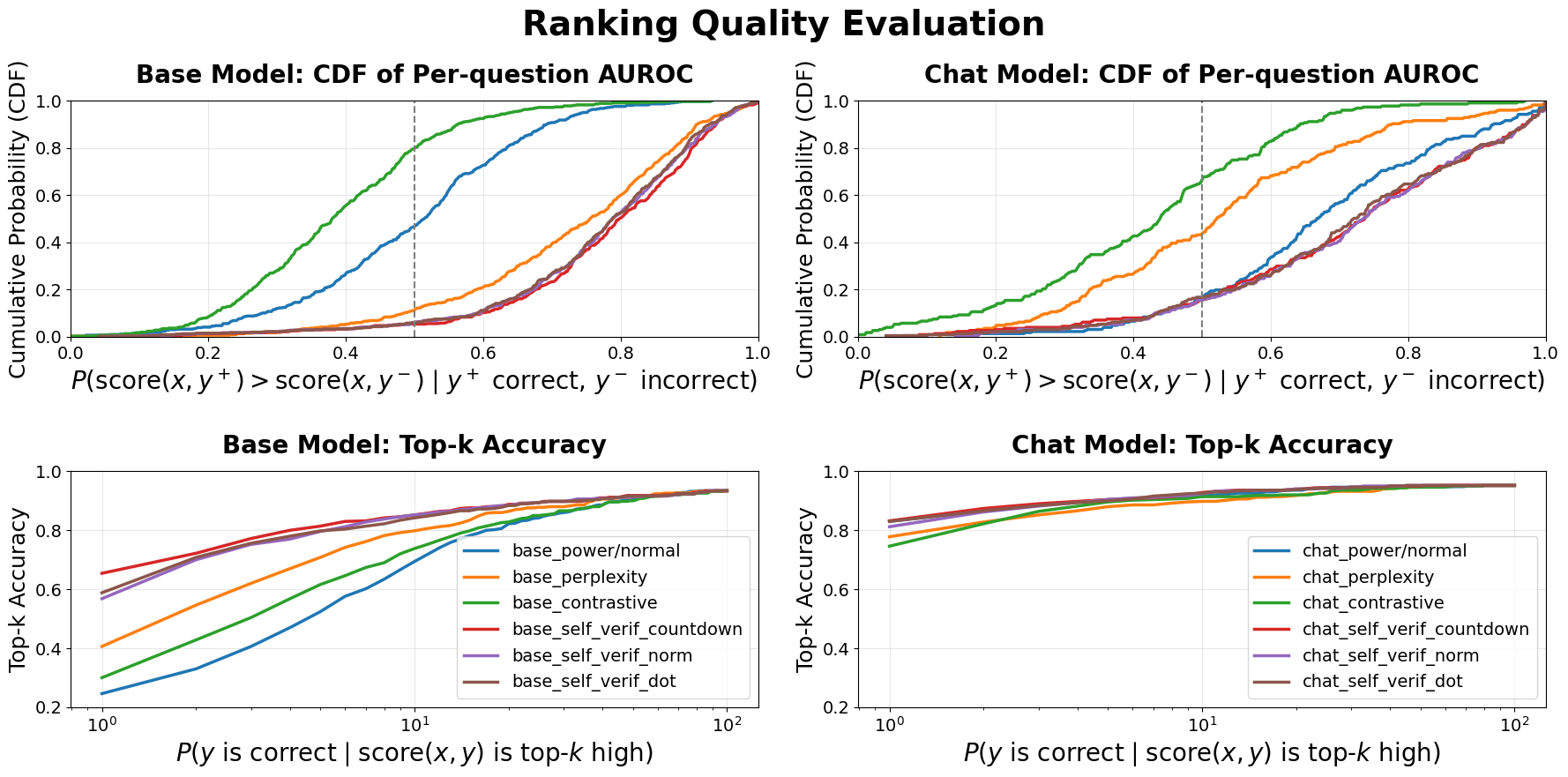}
    \caption{Full ranking quality of different scores for the Base (left) and Chat (right) models, including all questions (unfiltered). \textit{Top:} CDF of per-question AUROC over questions with valid pools. \textit{Bottom:} Top-$k$ accuracy on all questions.}
    \label{fig:ranking_quality_full} 
\end{figure*}

Figure~\ref{fig:ranking_quality_full} shows the full unfiltered version of Figure~1, 
including all questions regardless of oracle accuracy.

\section{Prompts}
\label{app:prompts}

\subsection{Proposal Prompt}
\label{app:prompt-proposal}

We sample candidate responses from $p_{\mathrm{LM}}$ using the following system prompt:

\begin{tcolorbox}[
  colback=gray!3,
  colframe=gray!50,
  boxrule=0.3pt,
  arc=2pt,
  left=4pt,right=4pt,top=4pt,bottom=4pt
]
\footnotesize
\ttfamily
\raggedright
You are a math assistant. Solve the problem step-by-step and provide your final answer in LaTeX format, ensuring the final result is placed inside \textbackslash boxed\{\}.
\end{tcolorbox}

\noindent For \textbf{base models}, the prompt is formatted as a plain string:

\begin{tcolorbox}[
  colback=gray!3,
  colframe=gray!50,
  boxrule=0.3pt,
  arc=2pt,
  left=4pt,right=4pt,top=4pt,bottom=4pt
]
\footnotesize
\ttfamily
\raggedright
\{system\_prompt\}\\[4pt]
Question: \{question\}\\[4pt]
Answer:
\end{tcolorbox}

\noindent For \textbf{chat models}, the prompt is formatted as a conversation:

\begin{tcolorbox}[
  colback=gray!3,
  colframe=gray!50,
  boxrule=0.3pt,
  arc=2pt,
  left=4pt,right=4pt,top=4pt,bottom=4pt
]
\footnotesize
\ttfamily
\raggedright
{[}\\[4pt]
\ \ \{\\[4pt]
\ \ \ \ "role": "system",\\[4pt]
\ \ \ \ "content": \{system\_prompt\}\\[4pt]
\ \ \},\\[4pt]
\ \ \{\\[4pt]
\ \ \ \ "role": "user",\\[4pt]
\ \ \ \ "content": \{question\}\\[4pt]
\ \ \}\\[4pt]
{]}
\end{tcolorbox}

\subsection{Self-Verification Prompt}
\label{app:prompt-selfverify}

For \textbf{base models}, the full verification prompt $v(x, y)$ has the following structure:

\begin{tcolorbox}[
  colback=gray!3,
  colframe=gray!50,
  boxrule=0.3pt,
  arc=2pt,
  left=4pt,right=4pt,top=4pt,bottom=4pt
]
\footnotesize
\ttfamily
\raggedright
\{Instruction\}\\[4pt]
\{Grading rules\}\\[4pt]
\#\#\# Problem: \{x\}\\[4pt]
\#\#\# Solution: \{y\}\\[4pt]
Take a moment to evaluate the logic internally. 
\{Virtual thinking filler\}\\[4pt]
\#\#\# Verdict:
\end{tcolorbox}

\noindent
For the \textbf{chat models}, the full $v(x,y)$ in formatted as a conversation:

\begin{tcolorbox}[
  colback=gray!3,
  colframe=gray!50,
  boxrule=0.3pt,
  arc=2pt,
  left=4pt,right=4pt,top=4pt,bottom=4pt
]
\footnotesize
\ttfamily
\raggedright
{[}\\[4pt]
\ \ \{\\[4pt]
\ \ \ \ "role": "user",\\[4pt]
\ \ \ \ "content": (\\[4pt]
\ \ \ \ \ \ "\{Instruction\}"\\[4pt]
\ \ \ \ \ \ "\{Grading rules\}"\\[4pt]
\ \ \ \ \ \ "\#\#\# Problem: \{x\}"\\[4pt]
\ \ \ \ \ \ "\#\#\# Solution: \{y\}"\\[4pt]
\ \ \ \ \ \ "Take a moment to evaluate the logic internally."\\[4pt]
\ \ \ \ )\\[4pt]
\ \ \},\\[4pt]
\ \ \{\\[4pt]
\ \ \ \ "role": "assistant",\\[4pt]
\ \ \ \ "reasoning\_content": "\{Virtual thinking filler\}",\\[4pt]
\ \ \ \ "content": "\#\#\# Verdict:"\\
\ \ \}\\
{]}
\end{tcolorbox}

\noindent
We add one sentence to elicit virtual thinking across all settings, including the no-filler (normal) condition.
By terminating the prompt with the incomplete string \texttt{``\#\#\# Verdict:''}, the model's next token is constrained to be \texttt{" ✅"}\ or \texttt{" ❌"}, whose log-probabilities feed directly into Eq.~\eqref{eq:score}.%
\footnote{In our implementation using the \texttt{qwen} tokenizer, the verdict tokens correspond to the two-token sequences \texttt{" ✅"} $\to$ \texttt{(25521, 227)} and \texttt{" ❌"} $\to$ \texttt{(51414, 234)}. We take $p_{\text{\tokC}}$ and $p_{\text{\tokI}}$ to be the probability of the \emph{first two} tokens in each sequence.}

\paragraph{Instruction.} The Instruction is as follows:
\begin{tcolorbox}[
  colback=gray!3,
  colframe=gray!50,
  boxrule=0.3pt,
  arc=2pt,
  left=4pt,right=4pt,top=4pt,bottom=4pt
]
\footnotesize
\ttfamily
\raggedright
Given the math problem, review the following solution carefully and determine if it is correct.\\
- If the solution is correct, output\\
\ \ ``\#\#\# Verdict: \tokC''.\\[2pt]
- If the solution contains a logical error, arithmetic error, or has wrong reasoning, output\\
\ \ ``\#\#\# Verdict: \tokI\\
\ \ \#\#\# Reason: \ldots''\\
\ \ with a brief explanation.\\[4pt]
\end{tcolorbox}

\paragraph{Grading Rules}
The grading rules for MCMC and candidate ranking are as follows:
\begin{tcolorbox}[
  colback=gray!3,
  colframe=gray!50,
  boxrule=0.3pt,
  arc=2pt,
  left=4pt,right=4pt,top=4pt,bottom=4pt
]
\footnotesize
\ttfamily
\raggedright
GRADING RULES:\\[2pt]
1. SCOPE: Judge both mathematical correctness and answer formatting. A solution must be mathematically sound AND produce a properly formatted final answer to be marked \tokC.\\[4pt]
2. FINAL ANSWER REQUIRED: A final answer MUST be present. It must appear inside \textbackslash boxed\{\} with no trailing punctuation or extra text inside the box. If no \textbackslash boxed\{\} answer is present, or the box contains extra text/punctuation, mark as \tokI.\\[4pt]
3. MATHEMATICAL CORRECTNESS: If the solution contains any logical error, arithmetic error, or invalid reasoning step that affects the final answer, mark as \tokI.\\[4pt]
4. NO PROGRESS / LOOPING: If the solution repeats the same step or sequence of steps three or more times without producing new intermediate results or progressing toward a final answer, mark as \tokI.\\[4pt]
5. NOTATION TOLERANCE: Minor notational or spacing differences (e.g.\ `1/2' vs `\textbackslash frac\{1\}\{2\}', extra whitespace, equivalent algebraic forms) are not errors.\\[4pt]
CRITICAL INSTRUCTION: Before outputting the `\#\#\# Verdict:' line, you must carefully think through your evaluation internally as the countdown progresses. Do not rush to give the answer---explicitly engage in silent, step-by-step reasoning during the countdown before revealing your conclusion.
\end{tcolorbox}

For SMC which requires scoring over partial solution sequence, the grading rules are as follows:
\begin{tcolorbox}[
  colback=gray!3,
  colframe=gray!50,
  boxrule=0.3pt,
  arc=2pt,
  left=4pt,right=4pt,top=4pt,bottom=4pt
]
\footnotesize
\ttfamily
\raggedright
GRADING RULES:\\[2pt]
1. A partial solution with no final answer yet should be judged ONLY on whether its steps and reasoning are correct so far. Do NOT penalize for being incomplete.\\[4pt]
2. If the solution is overly repetitive or cycling through the same steps without making progress, mark as \tokI.\\[4pt]
3. If the solution terminates early (e.g. with an end-of-text token) before any \\boxed{{}} answer appears, mark as \tokI as the model stopped without providing a final answer.\\[4pt]
4. If a final answer IS present, it must appear in \\boxed{{}} with no trailing punctuation or extra text inside the box, otherwise it is \tokI.\\[4pt]
5. Judge mathematical correctness only. Minor notational or spacing differences are not errors.\\[4pt]
CRITICAL INSTRUCTION: Before outputting the '\#\#\# Verdict:' line, you must carefully think through your evaluation internally as the countdown progresses. Do not rush to give the answer—explicitly engage in silent, step-by-step reasoning during the countdown before revealing your conclusion.
\end{tcolorbox}



\paragraph{Virtual thinking filler.} We investigate two virtual thinking variants:
\begin{itemize}
    \item Countdown filler.
    \begin{tcolorbox}[
      colback=gray!3,
      colframe=gray!50,
      boxrule=0.3pt,
      arc=2pt,
      left=4pt,right=4pt,top=4pt,bottom=4pt
    ]
    \footnotesize
    \ttfamily
    \raggedright
    Counting down: \$$n$, \$($n$-1), \ldots, 2, 1\ldots
    \end{tcolorbox}
    \noindent where $n$ is the countdown length. The decreasing integer sequence signals to the model that it should complete an internal reasoning process of commensurate depth before producing the verdict.
    
    \medskip
    \item Dot filler.
    \begin{tcolorbox}[
      colback=gray!3,
      colframe=gray!50,
      boxrule=0.3pt,
      arc=2pt,
      left=4pt,right=4pt,top=4pt,bottom=4pt
    ]
    \footnotesize
    \ttfamily
    \raggedright
    \ldots\,[$d$\,dots]\,\ldots
    \end{tcolorbox}
    \noindent where $d$ is the number of dot characters inserted. This variant is motivated by the observation that punctuation tokens can serve as implicit ``thinking'' tokens in autoregressive models~\citep[see, e.g.,][]{goyal2024think}, providing a lightweight inductive bias for deliberation without the semantic content of a countdown.
\end{itemize}
\noindent Neither variant requires the model to generate additional tokens; both are part of the input context and do not affect the single-forward-pass evaluation cost of $s_{\mathrm{sv}}$.

\subsection{Empirical Validation of Verdict Token Probabilities}
\label{app:score-validation}

We verify that the score $s_{\mathrm{sv}}$ in Eq.~\eqref{eq:score} is well-defined by confirming that the verdict tokens \tokC\ and \tokI\ together account for nearly all of the model's probability mass at the verdict position, i.e.\
\begin{equation}
    p_{\text{\tokC}}(x, y) + p_{\text{\tokI}}(x, y) \;\approx\; 1
    \label{eq:validation}
\end{equation}
holds empirically across inputs.

To validate Eq.~\eqref{eq:validation}, we compute $p_{\text{\tokC}}(x, y) + p_{\text{\tokI}}(x, y)$ for candidates $y$ in \textbf{Normal pool} using \textbf{Qwen3-1.7B}. 
Table~\ref{tab:verdict-validation} reports the mean, median, min, max, and 5th-percentile values of the summed probability across models on the Normal pool.
 
\begin{table*}[ht]
\centering
\begin{tabular}{llccccc}
\toprule
Model & Mean & Median & 5th pct. & Min & Max \\
\midrule
\textbf{Base} & 0.9588 & 0.9639 & 0.9209 & 0.71788& 1.0097\\
\textbf{Chat} & 0.9995 & 0.9995 & 0.9990 & 0.9395 & 1.0566 \\
\bottomrule
\end{tabular}
\caption{Summary statistics of $p_{\text{\tokC}} + p_{\text{\tokI}}$ across models on the Normal pool. Values close to 1.0 confirm the validity of the odds-ratio formulation. Values marginally exceeding 1.0 are due to floating-point rounding in the probability computation.}
\label{tab:verdict-validation}
\end{table*}

\subsection{Contrastive Decoding: Structured Analysis Prompt}
\label{app:prompt-contrastive}

For base models, the expert distribution in contrastive decoding is conditioned on a structured analysis prefix that decomposes the problem into type, tools, key quantities, traps, strategy, and a sanity check plan (truncated to 2,048 tokens).
The analysis is elicited via in-context learning using the following prompt:

\begin{tcolorbox}[
  colback=gray!3,
  colframe=gray!50,
  boxrule=0.3pt,
  arc=2pt,
  left=4pt,right=4pt,top=4pt,bottom=4pt
]
\footnotesize
\ttfamily
\raggedright
Q: If a train leaves Chicago at 9:00 AM traveling at 60 mph, and another train leaves New York (800 miles away) at 10:00 AM traveling at 80 mph, at what time will they meet?\\[4pt]
Analysis:\\
Problem Type \& Domain: This is a relative motion / meeting-point problem. Domain: elementary kinematics + algebra.\\
Mathematical Tools Required: linear equations, distance-rate-time formula (d = r$\cdot$t), system of two equations in one unknown.\\
Key Quantities \& Unknowns: Let t = hours after 9:00 AM. Train 1 distance: 60t. Train 2 distance: 80(t$-$1). Unknown: t where distances sum to 800.\\
Traps \& Edge Cases: Train 2 has a 1-hour head start offset --- forgetting this is the \#1 mistake. Units are consistent (miles, hours) --- no conversion needed.\\
Solution Strategy: Set up 60t + 80(t$-$1) = 800, solve for t, convert decimal hours to clock time.\\
Sanity Check Plan: Plug t back in, verify both distances sum to 800; check answer is after 10:00 AM.\\[4pt]
{[END OF ANALYSIS]}\\[4pt]
---\\[4pt]
Q: \{question\}\\[4pt]
Analysis: Problem Type \& Domain:
\end{tcolorbox}

\noindent The model's continuation of the \texttt{Analysis:} field constitutes the structured prefix; generation is truncated at 2,048 tokens before being prepended to the answer prompt.
\section{Failure Modes of Score-Guided Inference}
\label{sec:appendix-failure}

This appendix analyzes a fundamental failure mode shared by both the MCMC and SMC samplers introduced in Section~\ref{sec:sampling}, explains why the target distribution $p_{\mathrm{final}}(y \mid x) \propto p_{\mathrm{LM}}(y \mid x) \cdot s(y \mid x)$ is a principled choice in light of this failure mode, and derives the acceptance ratio under this target.

\subsection{Theoretical Analysis}
\paragraph{A unified diagnostic.}
Under our MCMC sampler with the prefix-resample proposal, a new candidate $y'$ is generated by sampling a shared prefix $y_{\mathrm{pre}}$ from the current response $y$ and continuing autoregressively. Conditioned on $y_{\mathrm{pre}}$, the proposal distributions are $q(y' \mid y, y_{\mathrm{pre}}) = p_{\mathrm{LM}}(y'_{\mathrm{suf}} \mid x, y_{\mathrm{pre}})$ and $q(y \mid y', y_{\mathrm{pre}}) = p_{\mathrm{LM}}(y_{\mathrm{suf}} \mid x, y_{\mathrm{pre}})$, where $y_{\mathrm{suf}}$, $y'_\mathrm{suf}$ are the suffices satisfying $y=y_{\mathrm{pre}}\|y_{\mathrm{suf}}$ and $y'=y_{\mathrm{pre}}\|y'_{\mathrm{suf}}$, with $\|$ denoting string concatenation. The resulting Metropolis--Hastings acceptance ratio is
\begin{align}
    \alpha
        &= \min\!\left(1,\; \frac{p_{\mathrm{LM}}(y_{\mathrm{suf}} \mid x, y_{\mathrm{pre}})\, p_{\mathrm{final}}(y' \mid x)}{p_{\mathrm{LM}}(y'_{\mathrm{suf}} \mid x, y_{\mathrm{pre}})\, p_{\mathrm{final}}(y \mid x)}\right) \notag \\
        &= \min\!\left(1,\; \frac{p_{\mathrm{LM}}(y \mid x)\, p_{\mathrm{final}}(y' \mid x)}{p_{\mathrm{LM}}(y' \mid x)\, p_{\mathrm{final}}(y \mid x)}\right) \notag \\
        &= \min\!\left(1,\; \frac{p_{\mathrm{final}}(y' \mid x) / p_{\mathrm{LM}}(y' \mid x)}{p_{\mathrm{final}}(y \mid x) / p_{\mathrm{LM}}(y \mid x)}\right), \label{eq:acceptance_rate}
\end{align}
where the second line follows from multiplying numerator and denominator by $p_{\mathrm{LM}}(y_{\mathrm{pre}} \mid x)$ and applying the chain rule $p_{\mathrm{LM}}(y_{\mathrm{pre}} \mid x)\,p_{\mathrm{LM}}(y_{\mathrm{suf}} \mid x, y_{\mathrm{pre}}) = p_{\mathrm{LM}}(y \mid x)$. If the target were $p_{\mathrm{final}} \propto s$ alone (without the $p_{\mathrm{LM}}$ factor), the acceptance ratio would then be
\begin{equation}
    \alpha=\min\!\left(1,\; \frac{s(y' \mid x) / p_{\mathrm{LM}}(y' \mid x)}{s(y \mid x) / p_{\mathrm{LM}}(y \mid x)}\right).
    \label{eq:mh-ratio}
\end{equation}
Similarly, the particle weights in our SMC sampler are proportional to $p_{\mathrm{final}}(y \mid x) / p_{\mathrm{LM}}(y \mid x)$; if the target were $p_{\mathrm{final}} \propto s$ alone, the importance weights would then be $s(y \mid x) / p_{\mathrm{LM}}(y \mid x)$.
In both cases, the quantity
\begin{equation}
    r(y \mid x) \;\coloneqq\; \frac{s(y \mid x)}{p_{\mathrm{LM}}(y \mid x)}
    \label{eq:diagnostic}
\end{equation}
is the central object governing sampler behavior: it appears as the ratio in the MCMC acceptance decision~\eqref{eq:mh-ratio}, and it is precisely the importance weight that would arise in SMC if $p_{\mathrm{LM}}$ were absent from the target.

\paragraph{Degeneration when the score is flat.}
Suppose $s(y \mid x)$ has low dynamic range relative to $p_{\mathrm{LM}}(y \mid x)$---that is, $s$ varies slowly across responses while $p_{\mathrm{LM}}$ varies sharply.
In this regime, the ratio in Eq.~\eqref{eq:mh-ratio} reduces to
\begin{equation}
    \alpha \approx \min\!\left(1,\; \frac{p_{\mathrm{LM}}(y \mid x)}{p_{\mathrm{LM}}(y' \mid x)}\right),
\end{equation}
and the score ceases to guide the search.
The same degeneration affects SMC---particle weights collapse to $1 / p_{\mathrm{LM}}(y \mid x)$, which favors longer responses regardless of their correctness, since $p_{\mathrm{LM}}$ assigns exponentially lower probability to longer sequences.
This reveals a shared failure mode: \emph{both MCMC and SMC fail to sample from $p_{\mathrm{final}}$ whenever the score lacks sufficient dynamic range relative to the language model.}

\paragraph{The power distribution as a structural fix.}
The only family of scoring functions that \emph{structurally} avoids the above failure mode for all inputs is the power distribution,
\begin{equation}
    p_{\mathrm{final}}(y \mid x) \;\propto\; p_{\mathrm{LM}}(y \mid x)^{\alpha},
    \label{eq:power}
\end{equation}
which corresponds to $s(y \mid x) = p_{\mathrm{LM}}(y \mid x)^{\alpha}$.
Under this choice, the diagnostic ratio~\eqref{eq:diagnostic} becomes $r(y \mid x) = p_{\mathrm{LM}}(y \mid x)^{\alpha - 1}$, which is itself a power of $p_{\mathrm{LM}}$ and hence always has well-matched dynamic range by construction.
No task-motivated score---including the verification score from Section~\ref{sec:score}---can provide this guarantee in general: although such scores are derived from $p_{\mathrm{LM}}$, the transformations involved (e.g., contrastive subtraction, self-verification prompting, virtual thinking countdown) are not monotone rescalings of $p_{\mathrm{LM}}$, and may therefore exhibit dynamic range mismatch on some inputs.

\paragraph{Pragmatic justification for $p_{\mathrm{final}} \propto p_{\mathrm{LM}} \cdot s$.}
While the power distribution offers a structural guarantee, it provides no mechanism for incorporating task-relevant signal beyond likelihood rescaling.
Our choice of $p_{\mathrm{final}}(y \mid x) \propto p_{\mathrm{LM}}(y \mid x) \cdot s(y \mid x)$ is motivated as the principled middle ground:
\begin{itemize}
    \item \textbf{Compared to $p_{\mathrm{final}} \propto s$ alone.} Including $p_{\mathrm{LM}}$ in the target ensures that the importance weights in SMC are proportional to $s(y \mid x)$ rather than $s(y \mid x) / p_{\mathrm{LM}}(y \mid x)$, directly suppressing the pathological $1/p_{\mathrm{LM}}$ term. The failure mode is reduced to cases where $s$ is genuinely uninformative---a property of the score itself, not an artifact of the framework.
    \item \textbf{Compared to $p_{\mathrm{final}} \propto p_{\mathrm{LM}}^\alpha$.} Multiplying by a task-motivated score $s$ injects correctness-oriented signal that temperature scaling alone cannot provide, enabling the sampler to concentrate on responses that are both fluent and factually grounded.
\end{itemize}
In summary, $p_{\mathrm{final}} \propto p_{\mathrm{LM}} \cdot s$ is the natural choice when $s$ is a task-motivated score: it inherits the dynamic-range benefits of anchoring to $p_{\mathrm{LM}}$ while still allowing the score to reshape the distribution toward correct responses.

\subsection{Empirical Illustration}
\label{sec:analysis-empirical}

To empirically verify the failure mode, we trace the MCMC chain and examine transitions where the score and the language model prior are in tension.
Concretely, for each proposed transition $(y \to y')$, we define
\begin{align}
    \Delta_s &= \log s(y' \mid x) - \log s(y \mid x), \\
    \Delta_\pi &= \log p_{\mathrm{LM}}(y' \mid x) - \log p_{\mathrm{LM}}(y \mid x).
\end{align}
Recall from Eq.~\eqref{eq:mh-ratio} that the log acceptance ratio is $\Delta_s - \Delta_\pi$.
A \emph{conflict} arises when $\Delta_s$ and $\Delta_\pi$ share the same sign: both the score and the prior nominally prefer the same candidate, yet because $p_{\mathrm{LM}}$ appears in the denominator of the acceptance ratio, it actively resists the transition that the score favors.
In such cases, the score \emph{wins} if its signal dominates, i.e., $|\Delta_s| > |\Delta_\pi|$, ensuring $\Delta_s-\Delta_\pi$ and $\Delta_s$ share the same sign: both prefer the same candidate.
We therefore compute two statistics over all conflicting transitions observed during the chain:
\begin{itemize}
    \item \textbf{Conflict rate}: the fraction of all proposed transitions that are conflicts,
    \item \textbf{Score win rate}: among conflicts, the fraction where $|\Delta_s| > |\Delta_\pi|$.
\end{itemize}
A high conflict rate paired with a low score win rate is the empirical signature of the failure mode: the prior dominates the acceptance decision, and the chain reduces to exploring $p_{\mathrm{LM}}$ rather than $p_{\mathrm{final}}$.

Table~\ref{tab:mcmc-conflict} reports these statistics for each scoring function considered in this work.

\begin{table*}[t]
    \begin{tabular}{lcc}
        \toprule
        \textbf{Score} & \textbf{Conflict Rate} & \textbf{Win Rate} \\
        \midrule
        Power Distribution Likelihood  & 1.0000 & 1.0000 \\
        Perplexity                     & 0.6538 & 0.0003 \\
        Contrastive Decoding           & 0.3004 & 0.3282 \\
        Self-Verification (ours)       & 0.4626 & 0.0411 \\
        \bottomrule
    \end{tabular}
    \centering
    \caption{
        Conflict rate and score win rate for each scoring function under the MCMC sampler.
        A high conflict rate with a low win rate indicates that $p_{\mathrm{LM}}$ dominates the acceptance decision.
    }
    \label{tab:mcmc-conflict}
\end{table*}

The results confirm the theoretical analysis.
The power distribution achieves a conflict rate and win rate of $1.0000$ by construction, serving as a sanity check: since $s(y \mid x) = p_{\mathrm{LM}}(y \mid x)^{\alpha}$, the diagnostic ratio becomes $r(y \mid x) = p_{\mathrm{LM}}(y \mid x)^{\alpha - 1}$, which amplifies the dynamic range of $p_{\mathrm{LM}}$ and ensures the score always dominates by algebraic guarantee.
Perplexity exhibits the most severe failure: despite a high conflict rate of $0.6538$, its win rate is effectively zero ($0.0003$), meaning the prior overwhelmingly dominates every contested transition.
This is expected: perplexity normalizes $\log p_{\mathrm{LM}}(y \mid x)$ by sequence length $|y|$, which \emph{flattens} the dynamic range of $p_{\mathrm{LM}}$ rather than amplifying it---the opposite effect to the power distribution---leaving $p_{\mathrm{LM}}$ in full control.
Self-verification, while exhibiting a more moderate conflict rate ($0.4626$), similarly suffers from a near-zero win rate ($0.0411$), indicating that its dynamic range remains insufficient relative to $p_{\mathrm{LM}}$ despite being a more informed score.
Contrastive decoding occupies a middle ground with a win rate of $0.3282$, suggesting that the contrastive formulation incidentally produces dynamic range more comparable to $p_{\mathrm{LM}}$---though still far from the structural guarantee of the power distribution.
Taken together, these results underscore that the failure mode is not merely a theoretical concern but a measurable phenomenon, and motivate the use of $p_{\mathrm{final}} \propto p_{\mathrm{LM}} \cdot s$ as a principled mitigation that anchors the target distribution to $p_{\mathrm{LM}}$.
 
\section{Self-Verification Prompt Design Ablations}
\label{app:virtual_ablation}
 
We ablate the self-verification (virtual-thinking) prompt along three
axes: the \emph{filler length}, the \emph{filler type} (a repeated dot
sequence vs.\ a descending countdown sequence), and---on the Chat
model---the \emph{placement} of the prompt within the chat template. We
further test whether any of these choices transfers to consensus
accuracy. All ablations use the same generation pool of $N{=}100$
candidates per question on MATH500, so that differences reflect the
scoring prompt alone and not variation in the candidate set. We report
three metrics. \textbf{Top-$k$ accuracy}---the fraction of questions for
which at least one correct candidate appears among the top-$k$
scored---captures precision-at-one quality relevant to best-of-$N$
selection. \textbf{Mean per-question Spearman correlation}---the Spearman
rank correlation between score and binary correctness computed within
each question and averaged over questions with valid pools (those
containing at least one correct and one incorrect candidate)---measures
overall within-question ordering. \textbf{Weighted-consensus
accuracy}---the fraction of questions whose score-weighted plurality
answer is correct---measures the downstream effect when the score is
used to reweight a majority vote rather than to select a single
candidate.
 
\subsection{Chat Model Virtual Thinking Behavior}
\label{app:chat_vt}
 
\paragraph{Settings.}
We examine how the Chat model responds to virtual thinking compared to the
Base model, using the same length sweep over
$n \in \{10, 20, 30, 40, 50, 60, 70, 80\}$ countdown tokens with the
no-filler baseline as reference (Figure~\ref{fig:ablation_length}, right
panel). We additionally compare dot and countdown fillers at matched
lengths to assess whether filler type interacts with instruction tuning.
 
\paragraph{Results.}
The Chat model behaves qualitatively differently from the Base model
under length variation. Whereas the Base model exhibits an approximate
unimodal pattern in Top-$k$ accuracy at small $k$, no consistent trend
emerges on the Chat model: curves at different lengths interleave without
a clear ordering, and mean per-question Spearman correlation likewise
shows no monotone relationship with $n$. Despite the absence of a
length trend, virtual thinking remains beneficial: the no-filler baseline
(\texttt{no\_count}) sits at or near the bottom of the Chat panel across
all $k$, confirming that inserting \emph{some} filler before the verdict
improves discrimination even when the optimal length is unclear. A second
difference from the Base model is that the dot filler occasionally
outperforms the countdown filler on Chat, whereas countdown consistently
leads on Base. We speculate that the Chat model, having been trained
on structured instruction-following data, is less sensitive to the
semantic content of the filler sequence and responds roughly equally to
any non-empty occupancy of the pre-verdict context window.
 
\begin{figure}[t]
  \centering
  \includegraphics[width=\columnwidth]{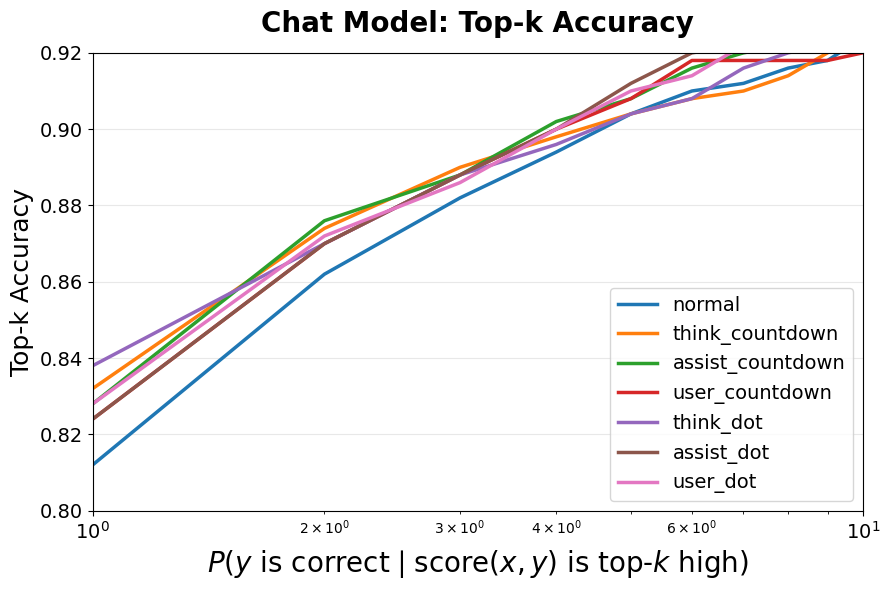}
  \caption{%
    \textbf{Effect of self-verification prompt placement on the Chat
    model.} Top-$k$ accuracy for the six placement$\times$filler-type
    variants---\{think, assist, user\}$\times$\{countdown, dot\}---against
    the default no-filler prompt (\texttt{normal}). Every placement lifts
    Top-$k$ accuracy at small $k$, but the gains are small and no
    placement is consistently best.}
  \label{fig:ablation_chat}
\end{figure}

\paragraph{Prompt placement.}
Because the Chat model uses a structured chat template, the
self-verification filler can be inserted in different turns: inside the
assistant's reasoning span (\emph{think}), as plain assistant text
(\emph{assist}), or in the user turn (\emph{user}). Figure~\ref{fig:ablation_chat}
sweeps these placements for both filler types. All six
placement$\times$type variants lift Top-$k$ accuracy at small $k$ over
the default no-filler prompt (\texttt{normal}), confirming once more
that some virtual thinking helps. But the gains are small, the curves
converge by moderate $k$, and no single placement dominates: the
\emph{think} placements are marginally ahead at $k{=}1$, yet the
ordering is not stable across $k$. Moreover, the placement variants
show no gain in mean per-question Spearman correlation over the default
prompt (Table~\ref{tab:sv_correlation}, bottom block), so the effect,
such as it is, is confined to precision-at-one. Placement thus behaves
like the other Chat-model design axes---a small, inconsistent
top-of-ranking effect with no change in overall ordering.
 
\subsection{Effect of Length}
\label{app:length}
 
\paragraph{Settings.}
We sweep the countdown filler length over
$n \in \{10, 20, 30, 40, 50, 60, 70, 80\}$ tokens, dot filler length of $d \in \{10, 20, 50\}$ with the no-filler
baseline (SV-none) included as a reference. Longer fillers give the model
more context-window occupancy before the verdict, potentially allowing a
more calibrated internal state, at the cost of additional compute per
scoring call. We report Top-$k$ accuracy as a function of $k$ on both
the Base and Chat models.
 
\paragraph{Results.}
 
\begin{figure*}[t]
  \centering
  \includegraphics[width=\textwidth]{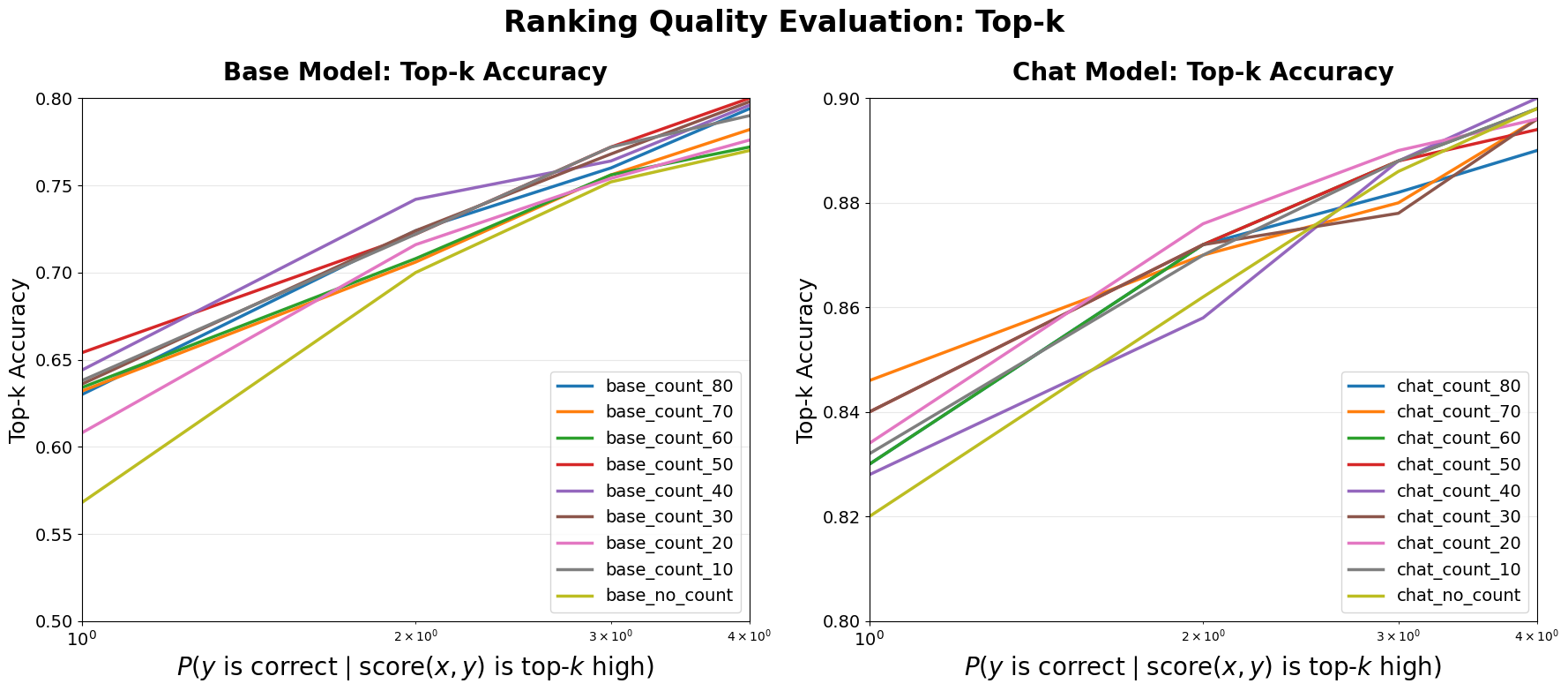}
  \caption{%
    \textbf{Effect of countdown filler length on Top-$k$ accuracy.}
    Each curve corresponds to one countdown length
    $L \in \{10,20,30,40,50,60,70,80\}$; the no-filler baseline
    (\texttt{no\_count}) is shown in yellow-green for reference.
    \emph{Left}: Base model. \emph{Right}: Chat model.
    On the Base model performance follows an approximate unimodal
    pattern as a function of $L$, peaking around $L{=}40$--$50$, with
    $L{=}10$ as a notable exception (see text). On the Chat model no
    consistent trend emerges.}
  \label{fig:ablation_length}
\end{figure*}
 
Figure~\ref{fig:ablation_length} shows Top-$k$ accuracy for each
countdown length on the Base and Chat models. On the Base model, the
relationship between length and performance is approximately unimodal
at small $k$: accuracy rises from the no-filler baseline and peaks
around $n{=}40$--$50$ before declining at longer lengths, consistent with
a deliberation-budget interpretation in which a moderate amount of
context occupancy sharpens the verdict without over-saturating the
model's attention. The exception is $n{=}10$, which at small $k$ performs
notably better than its position in the monotone region would
suggest---above $n{=}20$ and $n{=}30$. We attribute this to the frequency
of the countdown sequence ``$10, 9, \ldots, 1$'' in pretraining corpora,
where such sequences appear in everyday text (countdowns, lists, timers),
making this particular filler a familiar, semantically loaded context
rather than a neutral occupancy token; the model may therefore respond
to it with especially well-calibrated internal state. Both effects
diminish as $k$ grows and the curves converge, confirming that virtual
thinking's benefit is concentrated at the top of the ranking. On the
Chat model, no consistent ordering emerges across lengths: the curves
interleave throughout the $k$ range, suggesting that the
instruction-tuned model's output distribution is already
well-concentrated and less sensitive to the precise form of the
virtual-thinking filler.
 
Table~\ref{tab:sv_correlation} reports the mean per-question Spearman
correlation for all variants. On the Base model, all countdown lengths
improve substantially over the no-filler and norm baselines (${\approx}.355$),
clustering tightly between $.362$ and $.366$ with no monotone trend---the
spread across lengths is smaller than the gap to the baseline, suggesting
that \emph{any} countdown length is effective and the precise value matters
little for overall ranking quality. Dot fillers do not improve over the
no-filler baseline on Base and are marginally below norm. On the Chat
model, correlations are uniformly low (${\approx}.21$) across all variants
with no consistent ordering, consistent with the Top-$k$ finding.
 
\begin{table}
\centering
\begin{tabular}{l cc}
\toprule
\textbf{Variant} & \textbf{Base} & \textbf{Chat} \\
\midrule
\multicolumn{3}{l}{\textit{Baseline}} \\
No filler         & .3552 & .2115 \\
Norm              & .3552 & .2115 \\
\midrule
\multicolumn{3}{l}{\textit{Dot filler}} \\
Dot-10            & .3525 & .2121 \\
Dot-20            & .3559 & .2115 \\
Dot-50            & .3520 & .2106 \\
\midrule
\multicolumn{3}{l}{\textit{Countdown filler}} \\
Count-10          & .3654 & .2054 \\
Count-20          & .3645 & \textbf{.2141} \\
Count-30          & \textbf{.3659} & .2077 \\
Count-40          & .3643 & .2065 \\
Count-50          & .3650 & .2113 \\
Count-60          & .3625 & .2025 \\
Count-70          & .3639 & .2108 \\
Count-80          & .3652 & .2061 \\
\midrule
\multicolumn{3}{l}{\textit{Chat prompt placement (Chat only)}} \\
Normal            & ---   & .2133 \\
Think-countdown   & ---   & .2131 \\
Assist-countdown  & ---   & .2094 \\
User-countdown    & ---   & .2047 \\
Think-dot         & ---   & .2101 \\
Assist-dot        & ---   & .2091 \\
User-dot          & ---   & .2109 \\
\bottomrule
\end{tabular}
\caption{%
  \textbf{Mean per-question Spearman rank correlation} for all
  self-verification variants, grouped by filler type (baseline,
  dot, countdown) and, for the Chat model, by prompt placement.
  Bold marks the best value per column within the length-ablation
  groups. On the Base model, all countdown lengths yield a substantial
  improvement over the no-filler baseline with no clear ordering among
  lengths; dot fillers do not improve over baseline. On the Chat model,
  all variants cluster near $.21$ with no consistent trend.
  The prompt-placement block (bottom) is Chat-only; ``Think-countdown''
  is identical to Count-50.}
\label{tab:sv_correlation}
\end{table}
 
\begin{figure*}[t]
  \centering
  \includegraphics[width=\textwidth]{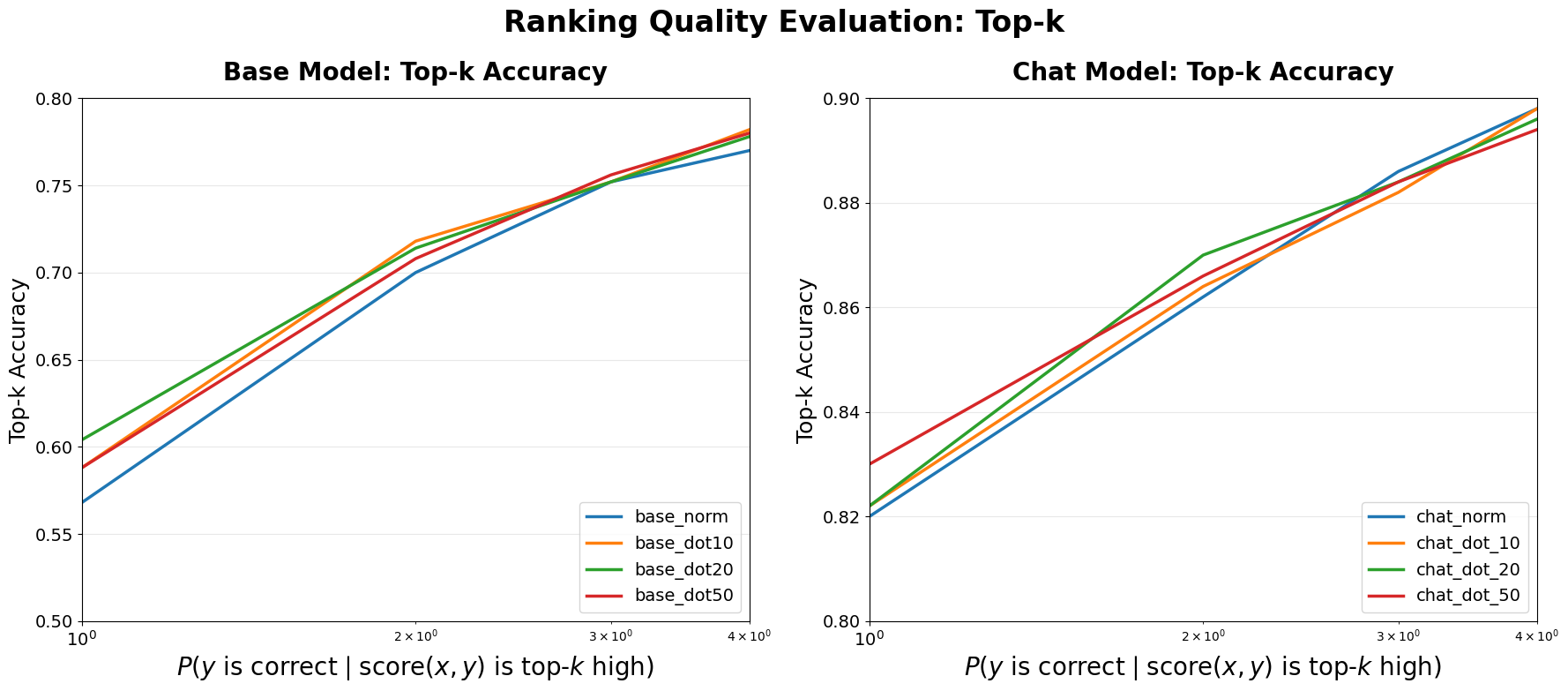}
  \caption{%
    \textbf{Effect of dot-filler length on Top-$k$ accuracy.}
    Length sweep for the dot filler, $L \in \{10,20,50\}$, with the
    no-filler baseline (\texttt{norm}) for reference.
    \emph{Left}: Base model. \emph{Right}: Chat model.
    Unlike the countdown filler (Figure~\ref{fig:ablation_length}), the
    dot filler produces only a small Top-$k$ separation from the
    baseline on the Base model and none on the Chat model, and shows no
    ordering by length.}
  \label{fig:ablation_dot}
\end{figure*}

\paragraph{Dot filler.}
Figure~\ref{fig:ablation_dot} repeats the length sweep for the dot
filler. On the Base model the dot curves lie modestly above the
no-filler baseline at small $k$, but the separation is smaller than for
the countdown filler (cf.\ Figure~\ref{fig:ablation_length}) and is not
ordered by length; in mean per-question Spearman correlation the dot
variants are statistically indistinguishable from the no-filler baseline
(Table~\ref{tab:sv_correlation}). The countdown filler thus improves
\emph{both} metrics, whereas the dot filler yields at most a weak
Top-$k$ effect and no gain in overall ordering---isolating the
\emph{semantic content} of the countdown sequence, rather than filler
length or mere context occupancy, as the active ingredient on the Base
model. On the Chat model the dot curves interleave with the baseline,
as for the countdown filler.
 
\paragraph{On adaptive filler length.}
The results above fix a single length for all questions. A natural
extension is a \emph{problem-adaptive} budget: map each question $x$ to
a filler length $L(x)$ based on an estimate of its difficulty, allocating
more deliberation to harder questions. One simple proxy is the oracle
accuracy of the candidate pool, though this is not available at test
time; a practical surrogate might be the entropy of the score distribution
over the pool. We leave a full evaluation of adaptive-length strategies
to future work.

\subsection{Effects on Consensus}
\label{app:vt_consensus}
 
\paragraph{Settings.}
The main text notes that virtual thinking barely affects consensus
accuracy---the dot and countdown fillers are near-indistinguishable when
$s_{\mathrm{sv}}$ is used as a reweighting signal for majority vote. Here
we confirm this finding across the full ablation grid. For every variant
we compute weighted-consensus accuracy: for each question we sum
$s_{\mathrm{sv}}$ over the candidates proposing each distinct answer and
take the highest-weighted answer as the consensus prediction. We report
this for the no-filler self-verification prompt, the dot filler at
$d\in\{10,20,50\}$, the countdown filler at $n\in\{10,\ldots,80\}$,
and---on the Chat model---the six placement variants. As a reference we
include \texttt{normal}, the unweighted (uniform-vote) majority baseline
that uses no self-verification signal; because all variants reweight the
same shared generation pool, \texttt{normal} is identical for the Base
and Chat scorers.
 
\paragraph{Results.}
 
\begin{figure}[t]
  \centering
  \includegraphics[width=0.92\columnwidth]{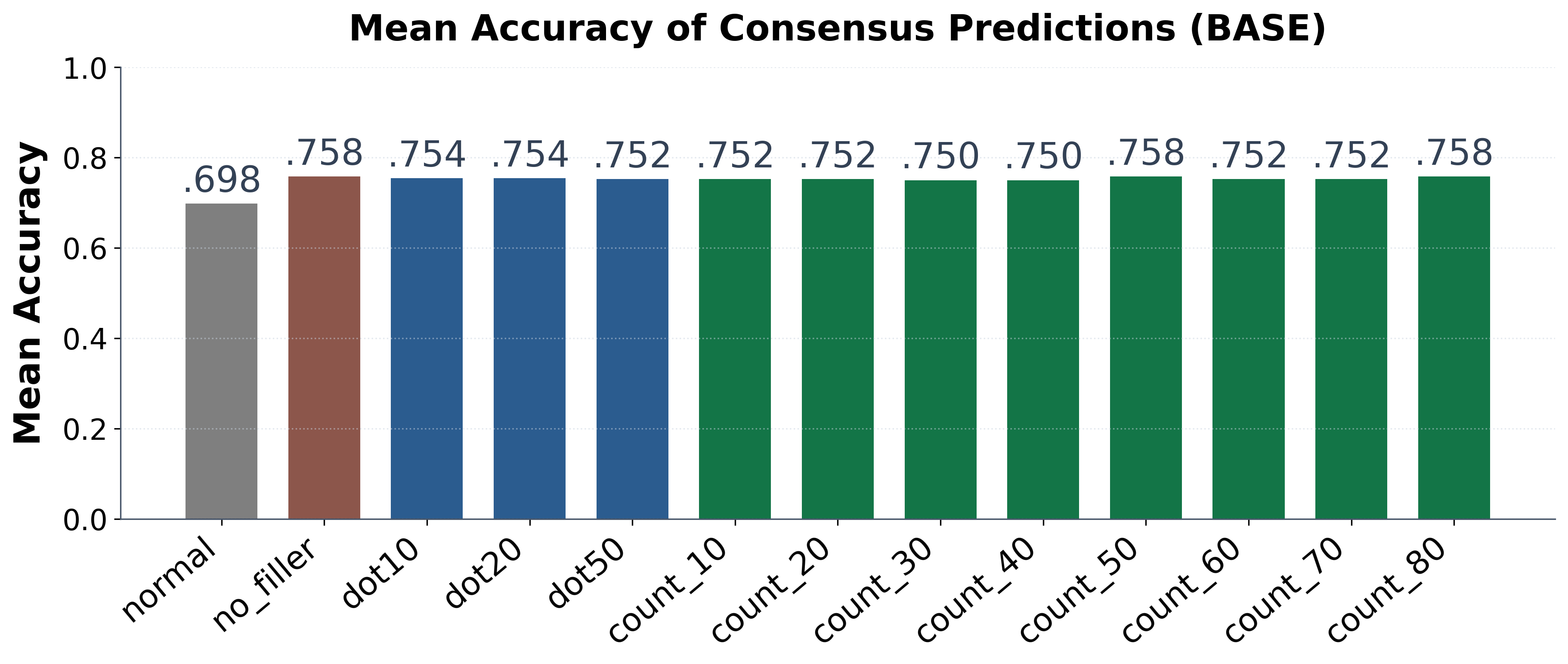}\\[4pt]
  \includegraphics[width=0.92\columnwidth]{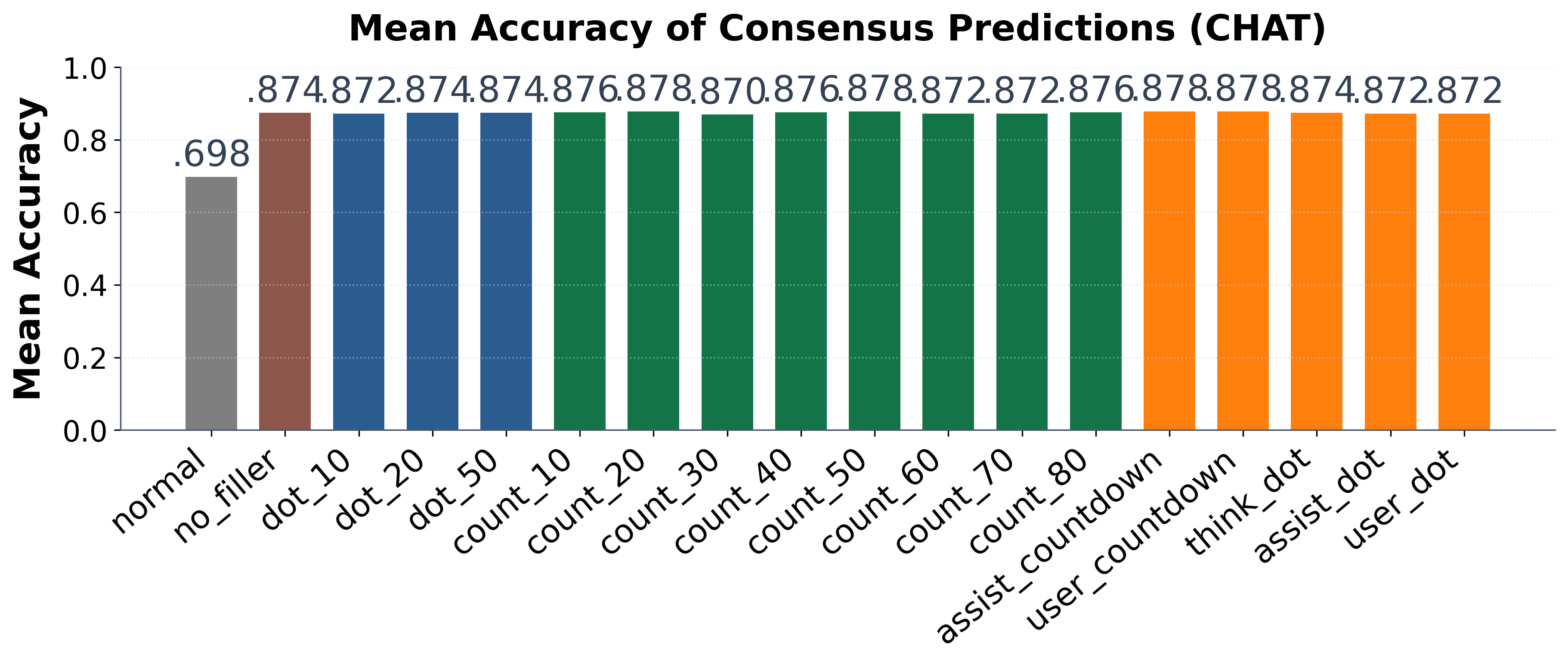}
  \caption{%
    \textbf{Weighted-consensus accuracy across all self-verification
    variants.} Each bar is the fraction of MATH500 questions whose
    score-weighted plurality answer is correct, over the shared
    $N{=}100$-candidate pool. \emph{Top}: Base scorer.
    \emph{Bottom}: Chat scorer. The leftmost bar (\texttt{normal}) is the
    unweighted majority-vote baseline and uses no self-verification
    signal. Reweighting by $s_{\mathrm{sv}}$ lifts consensus sharply
    ($.698\!\rightarrow\!{\approx}.75$ on Base,
    $.698\!\rightarrow\!{\approx}.87$ on Chat), but among the
    self-verification variants---across filler length, filler type, and
    prompt placement---accuracy varies by at most $.008$ on either
    model, with no consistent ordering.}
  \label{fig:vt_consensus}
\end{figure}
 
Figure~\ref{fig:vt_consensus} reports weighted-consensus accuracy for
every variant. The dominant effect is the presence of the
self-verification signal itself: reweighting the plurality vote by
$s_{\mathrm{sv}}$ raises consensus accuracy from the unweighted baseline
of $.698$ to ${\approx}.75$ on the Base scorer and to ${\approx}.87$ on
the Chat scorer. Conditioned on using a self-verification prompt,
however, none of the design axes moves the needle: the Base variants
span only $.750$--$.758$ and the Chat variants only $.870$--$.878$, a
range of $.008$ in both cases, and the ordering within that band is
inconsistent with the length, type, and placement trends seen for
ranking quality. In particular, the cheap no-filler prompt ($.758$ on
Base, $.874$ on Chat) already sits at the top of its cluster, tied with
the best countdown and placement variants.
 
Notably, the countdown filler's advantage on the Base model---clear in
Top-$k$ and Spearman---does \emph{not} carry over to consensus: the best
countdown variants ($.758$) only match the no-filler prompt. This is
expected. Consensus aggregates over the entire pool and is determined by
which \emph{answer} accumulates the most weight, so it is robust to
small perturbations in the score of any individual candidate; the gains
from a better filler are concentrated at the very top of the ranking,
exactly the regime that best-of-$N$ selection---but not plurality
voting---exploits. Self-verification prompt design is therefore a
selection-time lever, not a consensus-time one.

\subsection{Summary}
\label{app:vt_summary}
 
Across all three ablation axes the picture is consistent. The decision
that matters is binary---whether to use a self-verification signal at
all. Adding one produces large gains in every metric (Top-$k$, Spearman,
and consensus) over the \texttt{normal} baseline; conditioned on using
one, the prompt's exact form is close to a free parameter. No design
choice yields a statistically meaningful difference in consensus
accuracy (all variants fall within a $.008$ band), and on the Chat model
no choice yields a meaningful difference in ranking quality
either---Top-$k$, Spearman, and consensus curves all interleave without
a stable order.
 
The one real---if modest---difference appears in best-of-$N$
\emph{selection} on the Base model. There the countdown filler improves
both Top-$k$ accuracy at small $k$ and mean per-question Spearman
correlation (${\approx}.365$ vs.\ ${\approx}.355$ for the no-filler and
dot prompts), consistently across all swept lengths, whereas the dot
filler does not improve overall ordering. This advantage is specific to
the top of the ranking and does not transfer to consensus, where every
variant is tied.
 
We draw three design recommendations. For best-of-$N$ selection with a
Base-style model, use a countdown filler; the precise length is not
critical, as any value in the swept range works, with a mild preference
for $n{=}40$--$50$. For a Chat-style instruction-tuned model the filler
form is immaterial---dot, countdown, and placement are
interchangeable---so the cheapest option, a short or no-filler prompt,
is preferred. And when the downstream goal is consensus or weighted
majority vote rather than single-candidate selection, self-verification
prompt design can be ignored entirely: all variants, including the
inexpensive no-filler prompt, are statistically indistinguishable, so no
tuning is warranted.
 
\section{Scoring Functions as Confidence for Abstention}
\label{app:calibration}
  \begin{table*}[!ht]
\centering
\begin{tabular}{l rrrr rrrr}
\toprule
 & \multicolumn{4}{c}{\textbf{Base}} & \multicolumn{4}{c}{\textbf{Chat}} \\
\cmidrule(lr){2-5}\cmidrule(lr){6-9}
\textbf{Score} & All & {$<$.90} & {$<$.80} & {$<$.50}
               & All & {$<$.90} & {$<$.80} & {$<$.50} \\
\midrule
Power           &  .338 &  .305 &  .247 &  .111 & \textbf{.493} &  .301 &  .236 &  .154 \\
Perplexity      &  .252 &  .250 &  .239 &  .123 & $-$.148       & $-$.119 & $-$.042 & $-$.012 \\
Contrastive     &  .036 &  .006 & $-$.024 & $-$.042 & $-$.323   & $-$.091 & $-$.032 & $-$.020 \\
SV-countdown    & \textbf{.510} & \textbf{.497} & \textbf{.443} & \textbf{.275}
                &  .472 &  .323 &  .280 &  .217 \\
SV-norm         &  .483 &  .474 &  .425 &  .267 &  .474 &  .312 &  .274 & \textbf{.219} \\
SV-dot          &  .489 &  .479 &  .428 &  .266 &  .479 & \textbf{.325} & \textbf{.281} &  .218 \\
\bottomrule
\end{tabular}%
\caption{%
  \textbf{Spearman rank correlation with correctness} under four
  difficulty thresholds (``All'' = no filter; $<$0.90/$<$0.80/$<$0.50
  retain questions whose oracle accuracy falls below the threshold).
  Bold marks the best score per column.
  Negative values indicate anti-correlation.
  Chat-power leads on easy questions but collapses under thresholding;
  self-verification variants are the most robust across difficulty levels.}
\label{tab:spearman-corr}
\end{table*}
Ranking candidates \emph{within} a question and deciding \emph{whether to
answer at all} make different demands on a score. Ranking needs only the
correct \emph{ordering} of candidates inside each question; abstention
needs more---scores must be comparable \emph{across} questions, so that a
single global threshold can separate the questions a model should answer
from those it should decline. Whether a score meets this stronger
requirement depends on its \emph{calibration scope}: the range over which
its magnitudes carry a consistent meaning. Calibration scope is set by the
score's functional form, not by model quality, and it differs from one
scoring function to the next. We therefore study all four scoring
functions side by side as confidence signals for abstention, asking how
far each one's calibration scope extends and what limits it.
 
\paragraph{Settings.}
We pool all $N{\times}|\mathcal{X}|$ (question, candidate) pairs and
report the \textbf{Spearman rank correlation} between the score
$s(y \mid x)$ and binary correctness $c(y)$ over the pooled pairs.
Pooling is what turns this into a test of calibration scope: a score that
orders candidates well \emph{within} each question but whose magnitudes
are not comparable across questions will rank a confident wrong candidate
from one question above a less-confident correct candidate from another,
depressing the pooled correlation. Because Spearman depends only on the
score ordering, it is invariant to any monotone rescaling and places all
four scoring functions on equal footing. To probe difficulty-conditioned
discrimination we recompute it on subsets filtered by per-question oracle
accuracy: the ``$<$0.90'', ``$<$0.80'', and ``$<$0.50'' columns
progressively retain only harder questions, whose candidate pools contain
fewer correct answers and on which an abstention decision is most
consequential.

\paragraph{Results.}
Table~\ref{tab:spearman-corr} reports the Spearman rank correlation under
four difficulty thresholds, and the scores separate cleanly by
calibration scope.

Self-verification has the widest scope. $s_{\mathrm{sv}}$ is the
normalized probability of a single verdict token---an estimate of
$P(\text{correct})$ on a fixed $[0,1]$ scale, computed identically for
every question and free of the mechanical length penalty that afflicts
likelihood products. Its magnitudes are comparable across questions by
construction, and this is visible as robustness: SV-countdown leads every
column on the Base model, and all three SV variants retain a substantial
pooled correlation (${\approx}.27$) even on the hardest subset ($<$0.50),
after the easy questions that inflate other scores have been filtered
away.
 
The likelihood-derived scores are calibrated, at best, only \emph{within}
a question, and each is bounded by a different artifact of its functional
form. \textbf{Power}, $p_{\mathrm{LM}}(y\mid x)^{\beta}$, applies no
length normalization: sequence probability decays geometrically in $|y|$
and the exponent $\beta{=}4$ sharpens the decay, so power rewards short
responses regardless of correctness. Pooled across questions this length
bias dominates---power's high unfiltered correlation on the Chat model
($.493$) is borrowed from the easy regime, where the model is confident
and its probability mass genuinely tracks correctness; once easy
questions are removed, the length and predictability confound is all that
remains and the correlation collapses to $.236$ at $<$0.80 and $.154$ at
$<$0.50, below every SV variant. \textbf{Perplexity},
$p_{\mathrm{LM}}(y\mid x)^{1/|y|}$, is length-normalized and so escapes
the short-response bias, but the quantity it measures---average per-token
predictability---is fluency, not correctness; across questions a generic
or formulaic answer scores high whether or not it is right. This leaves a
weak positive signal on the Base model and an actively misleading one on
the Chat model, where the pooled correlation is negative throughout
($-.148$ on ``All'') because the instruction-tuned model emits fluent,
confident-sounding wrong answers that perplexity rates highly.
\textbf{Contrastive}, the expert/amateur likelihood ratio, cancels length
to first order---both terms are products over the same $|y|$
tokens---but its magnitude tracks the expert/amateur gap, which varies
per question and, on the Base model, depends on an engineered analysis
prefix; it carries almost no pooled signal on Base ($.036$) and an
anti-correlated one on Chat ($-.323$). The pattern is consistent: only
self-verification has a calibration scope wide enough for cross-question
abstention, while the likelihood scores are within-question instruments
whose magnitudes, read across questions, import whichever
artifact---response length, fluency, or a question-dependent
baseline---their functional form fails to cancel.

\end{document}